\documentclass[letterpaper]{article} % DO NOT CHANGE THIS
\usepackage{aaai25}  % DO NOT CHANGE THIS
\usepackage{times}  % DO NOT CHANGE THIS
\usepackage{helvet}  % DO NOT CHANGE THIS
\usepackage{courier}  % DO NOT CHANGE THIS
\usepackage[hyphens]{url}  % DO NOT CHANGE THIS
\usepackage{amsmath}
\usepackage{tikz}
\usepackage{listings}
\usepackage{lstautogobble}  %
\usepackage{color}          %
\usepackage{zi4}            %
\definecolor{bluekeywords}{rgb}{0.13, 0.13, 1}
\definecolor{greencomments}{rgb}{0, 0.5, 0}
\definecolor{redstrings}{rgb}{0.9, 0, 0}
\definecolor{graynumbers}{rgb}{0.5, 0.5, 0.5}
\usetikzlibrary{spy}
\usepackage{amssymb}
\usepackage{graphicx} % DO NOT CHANGE THIS
\usepackage{natbib}  % DO NOT CHANGE THIS AND DO NOT ADD ANY OPTIONS TO IT
\usepackage{caption} % DO NOT CHANGE THIS AND DO NOT ADD ANY OPTIONS TO IT
\usepackage{algorithm}
\usepackage{algpseudocode}
\usepackage{comment}
\usepackage{listings}

\usepackage{newfloat}
\usepackage{listings}
\DeclareCaptionStyle{ruled}{labelfont=normalfont,labelsep=colon,strut=off} % DO NOT CHANGE THIS
\floatstyle{ruled}
\newfloat{listing}{tb}{lst}{}
\floatname{listing}{Listing}
 \nocopyright %-- Your paper will not be published if you use this command
\title{MVGaussian: High-Fidelity text-to-3D Content Generation with Multi-View Guidance and Surface Densification}
\author{
    Phu Pham \textsuperscript{\rm 1} \equalcontrib, Aradhya N. Mathur \textsuperscript{\rm 2} \equalcontrib, Ojaswa Sharma \textsuperscript{\rm 2}, Aniket Bera \textsuperscript{\rm 1}
}

\affiliations{
    \textsuperscript{\rm 1}Purdue University \\ \textsuperscript{\rm 2}IIITD  \\
    phupham@purdue.edu, aradhya@iiitd.ac.in, ojaswa@iiitd.ac.in, aniketbera@purdue.edu

}
\usepackage{bibentry}
\begin{document}

\maketitle

\begin{abstract}

The field of text-to-3D content generation has made significant progress in generating realistic 3D objects, with existing methodologies like Score Distillation Sampling (SDS) offering promising guidance. However, these methods often encounter the ``Janus" problem—multi-face ambiguities due to imprecise guidance. Additionally, while recent advancements in 3D gaussian splatting have shown its efficacy in representing 3D volumes, optimization of this representation remains largely unexplored. This paper introduces a unified framework for text-to-3D content generation that addresses these critical gaps. Our approach utilizes multi-view guidance to iteratively form the structure of the 3D model, progressively enhancing detail and accuracy. We also introduce a novel densification algorithm that aligns gaussians close to the surface, optimizing the structural integrity and fidelity of the generated models. Extensive experiments validate our approach, demonstrating that it produces high-quality visual outputs with minimal time cost. Notably, our method achieves high-quality results within half an hour of training, offering a substantial efficiency gain over most existing methods, which require hours of training time to achieve comparable results. 
% Project page: \textit{https://mvgaussian.github.io/}.

\end{abstract}
% \footnote{* represents equal contribution}
\begin{links}
    \link{Code}{https://mvgaussian.github.io}
    % \link{Datasets}{https://aaai.org/example/datasets}
    % \link{Extended version}{https://aaai.org/example/extended-version}
\end{links}

\section{Introduction}
Recent advancements in text-to-3D generation have opened new avenues for creating complex 3D content directly from textual descriptions. This capability is crucial as it provides a straightforward, intuitive means for creators across various industries like gaming, virtual reality, and film-making, enabling rapid prototyping and visualization without the need for advanced modeling software or specialized training.

In leveraging foundation models for image generation, recent works have used reconstruction methods like Neural Radiance Fields (NeRFs) \cite{NERF} and 3D Gaussian Splatting (3DGS) \cite{3DGS} to make significant strides in the field. These models typically utilize Score Distillation Sampling (SDS) \cite{dreamfusion} to train the NeRF or Gaussian splatting methods, allowing for the generation of consistent 3D representations suitable for high-quality rendering and mesh extraction.

Recent approaches, \cite{gsgen, luciddreamer, prolificdreamer, gaussiandreamer, dreamgaussian}, despite successfully generating 3D models, encounter significant challenges that limit their practical application. These challenges include the multi-face (or Janus) problem, where models produce outputs with inconsistent appearances when viewed from different angles, lengthy training times, and a general lack of fine detail in the generated models. Furthermore, most methods suffer from issues related to the complexity of their moving parts and hyperparameters. They require considerable computational resources and time to generate high-quality content, or they compromise on quality to achieve faster processing times due to the inherent quality-time tradeoff.
%YOU NEED TO SHOW THESE RESULTS IN A TABLE (COMPUTATION TIME VS ACCURACY ETC.)

% TODO: explain more about the method
To address these limitations, we propose a novel framework that enhances the text-to-3D transformation process by integrating SDS with an efficient 3D Gaussian splatting technique for robust 3D representation. Our approach not only addresses the aforementioned issues but also significantly reduces the computational overhead and time required for training. Our contributions can be summarized as follows:

%%%%%
%Foundation models for image generation have been leveraged lately with a combination of reconstruction methods such as NeRFs and 3DGS allowing for unprecedented progress in the field of text to 3D generation. Stable diffusion methods have been employed to train the NeRF/ Gaussian splatting methods using Score Distillation Sampling techniques to generate consistent 3D representation which can be used for high-quality rendering and mesh extraction. However, most of these methods suffer from different limitations due to a lot moving part and hyperparameters. Most of these methods require considerable time and compute for the generation of high-quality renders or are unable to produce good quality if they are fast due to a quality-time tradeoff. We aim to bridge this gap by improving the densification scheme which seems to have a tremendous impact on the generation of high-qaulity renders. Since, the stable diffusion pipeline is fixed, the densification and pruning component of the 3DGS in most methods forms the bottleneck for high quality generation along with the time taken for the training process.  
 
\begin{itemize}
    \item We introduce a unified framework for text-to-3D content generation that integrates SDS loss with 3D Gaussian splatting. This integration enables the rapid generation of high-fidelity 3D content while effectively addressing the multi-face problem by maintaining consistency across different viewpoints.
    
    \item We propose a novel densification method by optimizing the placement and density of gaussian elements that accelerate the generation process. We reduce the overall training time to only 25 minutes while still achieving high-quality results. This method aims to bridge the gap in the densification scheme, which significantly impacts the quality of renders. 
    % COMMENT: [30 MINS IS RELATIVE, PLEASE GIVE SOME COMPARISON WITH SOTA HERE, OR PERHAPS SAY X\% LESSER THAN SOTA ]

    \item Through rigorous experiments, we demonstrate that our method not only matches but often surpasses the quality of existing approaches with shorter training time. These experiments validate the effectiveness of our approach in producing detailed and realistic 3D models from textual descriptions. 
    % COMMENT: MENTIONS SOME METRICS HERE AND IMPROVEMENT NUMBERS

\end{itemize}

\section{Related work}

Recent advancements in text-to-3D generation have built on the foundations established by text-to-image generation, 3D representations, and techniques for lifting 2D images to 3D models. This section reviews significant contributions in these areas, highlighting their methodologies and addressing their limitations.

\subsection{Text-to-image generation}

Diffusion models \cite{ho2020denoising, song2020denoising} have significantly advanced high-quality generative modeling, accelerating the field of content generation. Stable diffusion \cite{rombach2022high} has demonstrated the effectiveness of diffusion over latent spaces for producing high-quality conditioned generation, particularly for text-to-image tasks. Methods like DALL-E \cite{ramesh2021zero} and Imagen \cite{imagen} leverage text embeddings, such as CLIP \cite{radford2021learning}, to jointly train text and image encoding and decoding. These models are trained on large datasets, enabling zero-shot image generation.

% \subsection{3D representations}
% Recently, researchers \cite{NERF} have made significant advancements in 3D rendering through the development of simple and fast deep learning-based methods that utilize shallow MLPs (Multi-Layer Perceptrons) to learn radiance fields for given scenes. These techniques employ shallow MLPs to construct a neural radiance field, leveraging volume rendering techniques. The MLP predicts the $RGB\sigma$ values based on ray information generated from a camera within the scene, enabling it to render the corresponding view. The MLPs are trained on a set of views from a given scene, facilitating novel view synthesis. This approach has proven effective for sparse views and has been extended to temporal scenes as well \cite{cao2023hexplane}.

% Additionally, \cite{3DGS} introduced a shift from implicit representations like NeRF to explicit representations using 3D gaussians. These gaussians are rendered using a differentiable tile rasterization technique to achieve the target view. The method optimizes the position and covariance of the gaussians by decomposing them into scale, rotation, opacity, and color. Furthermore, gradient-based densification and pruning schemes are proposed to manage the automatic addition and removal of gaussians in a scene. This approach allows for fast, real-time radiance field rendering, offering significant improvements in speed and efficiency over NeRF-based methods. As a result, this method has largely supplanted NeRFs in many applications due to its superior performance in neural rendering.

\subsection{3D Representations}
Recent advancements in 3D rendering have focused on using shallow MLPs to learn radiance fields for scenes, leveraging volume rendering techniques. These MLPs predict $RGB\sigma$ values from camera ray data to render views, proving effective for novel view synthesis even with sparse views. This approach has also been extended to temporal scenes \cite{NERF, cao2023hexplane}.

Shifting from implicit to explicit representations, \cite{3DGS} introduced 3D gaussians with a differentiable rasterization technique for faster, real-time rendering. This method optimizes gaussian parameters like scale, rotation, opacity, and color, and includes gradient-based schemes for managing gaussians in a scene. Due to its speed and efficiency, this approach has largely replaced NeRFs in many applications.

\subsection{Lifting to 3D}
Building on previous methods, \citet{dreamfusion} introduced a novel approach for generating 3D models by leveraging text-to-image models. They use a pre-trained diffusion model to distill view information into NeRF models. The Score Distillation Sampling (SDS) technique, central to this process, optimizes the NeRF model by omitting the U-Net Jacobian term. However, limitations like blurry renderings and the Janus problem arise due to the lack of 3D awareness in the Stable Diffusion model.

To address these, \citet{wang2023score} introduced voxel radiance fields and Score Jacobian Chaining, improving image quality but still facing the Janus problem and flatness. Prolific Dreamer \cite{prolificdreamer} then proposed Variational Score Distillation, which enhances visual quality, diversity, and robustness.

DreamGaussian \cite{dreamgaussian} replaced the NeRF-based representation with 3DGS, achieving faster text-to-3D and image-to-3D generation but still encountering issues like the Janus problem and poor mesh quality. Subsequent methods like GSGen \cite{gsgen}, LucidDreamer \cite{luciddreamer}, and GaussianDreamer \cite{gaussiandreamer} aimed to improve by using Point-e \cite{nichol2022point} for initialization, but bottlenecks persist as Point-e does not generalize for complex prompts.

In contrast, \citeauthor{chen2023fantasia3d} \shortcite{chen2023fantasia3d} offered a unique approach by disentangling geometry and appearance for high-quality 3D generation, using a DMTet-based hybrid surface representation \cite{shen2021deep}.

%They employ DMTet-based hybrid surface representation \cite{shen2021deep} for geometry modeling and learn an MLP to predict the SDF value and position offset for the tetrahedral grid of DMTet. Then use the Physically Based Rendering material model and use SDS loss for computing the material-based rendering.    
% Dreamgaussian: poor quality due to coarse gaussians and inefficient mesh extraction, 2-stage approach. Fast to train but huge tradeoff for quality. 

%ProlificDreamer: slow, janus problem

% Fanatasia3D and a few other methods: slow due to NERF?

% Gsgen: high quality, the texture is good but relies on point\_e for intialization, Point\_e cannot handle complex prompt and thus can easily fail if the initialization is not good enough. Need to find a suitable point\_e prompt.

% LucidDreamer: high quality, but again relies on point\_e initialization. Utilizing different diffusion models (civitai checkpoints)

% All the above methods suffer from multiface (janus) problem due to the ambiguitity of the 2D diffusion guidance.

% ================ \\
% CREATE a montage to show case the weaknesses of these approaches.

% MVDream: fine-tune diffusion model using 3D data. Gives good guidance. However, it slows to train (need to check the running time).

% Identify GAPs: we aim to generate high-fidelity 3D content from text with less training time
% THIS ENTIRE SECTION BELOW HAS A LOT OF PREVIOUS STUFF WHICH ISN'T DIRECTLY RELEVANT, CAN YOU ONLY INTRODUCE THE PARTS WHICH WE NEED AND THEN REFER TO THE PAPER, YOU CAN EDIT THIS AT THE END BASED ON HOW MUCH SPACE YOU NEED.
% \section{Preliminaries}
\section{Background}
\subsection{Diffusion process}

Diffusion has emerged as a pivotal approach in generative modeling, particularly for text-to-image generation tasks \cite{ramesh2022hierarchical, zhang2023adding, imagen}. Recent advancements show that diffusion models not only surpass traditional generative adversarial networks (GANs) \cite{GAN} in image quality but also provide improved training stability and convergence \cite{DiffvsGAN, DDPM}. These models simulate the reverse process of diffusion, starting with corrupted input data and progressively reconstructing it back to the original form. In text-to-image applications, the diffusion process is typically applied in the latent space, which reduces dimensionality, accelerates computations, and lowers memory requirements while preserving essential data features for high-quality generation.

\subsection{Score Distillation Sampling}

The Score Distillation Sampling (SDS) \cite{dreamfusion} is an optimization strategy that integrates the generative capability of pre-trained 2D diffusion models into the synthesis of 3D objects. This is achieved by optimizing a 3D model such that its 2D projections, when viewed from random views, mimic the image distribution learned by a diffusion model. The key aspect of SDS is its utilization of score functions derived from the diffusion process to inform the optimization, thereby ensuring that the generated 3D models align with the statistics of images conditioned on the text.

% The SDS approach minimizes a loss derived from the Kullback-Leibler (KL) divergence, measuring the discrepancy between the distribution of the rendered 2D images of the 3D model and the distribution modeled by the diffusion process. This loss ensures that the generated images align with the trajectories learned by the diffusion model:
% \begin{equation}
%     L_{\text{SDS}} = \mathbb{E}_t \left[ \text{KL}(q(\mathbf{z}_t | g(\theta)) \| p_{\phi}(\mathbf{z}_t)) \right]    
% \end{equation}

% where \( g \) represents the renderer of the 3D model parameterized by \( \theta \). 

The score functions \( s_{\phi}(\mathbf{z}_t; \theta) = -\epsilon_{\phi}(\mathbf{z}_t; \theta) / \sigma_t \) derived from the diffusion model are used to compute updates for \( \theta \). These functions provide gradients of the log probability with respect to the latents, guiding the optimization to areas of higher probability under the model's distribution.

%\[ s(\mathbf{z}_t; \theta) = \nabla_{\mathbf{z}_t} \log p(\mathbf{z}_t; \theta). \]

Then, as shown by \citet{dreamfusion}, the gradient of the \( L_{\text{SDS}}\) can be computed as:
\begin{equation}
    \nabla_{\theta}L_{\text{SDS}} = \mathbb{E}_{t,\epsilon} \left[ w(t) (\hat{\epsilon}_{\phi}(\mathbf{z}_t) - \epsilon) \frac{\partial \mathbf{x}}{\partial \theta} \right]
\end{equation}

where \( \hat{\epsilon}_{\phi} \) is the noise prediction by the model, \( w(t) \) is a weighting function depending on the time step \( t \), \( \epsilon \) is the noise added in the forward process, \( \mathbf{x} \) is the rendered image of the 3D model represented by \( \theta \).

\begin{figure*}
    \centering
    \includegraphics[width=0.9\linewidth]{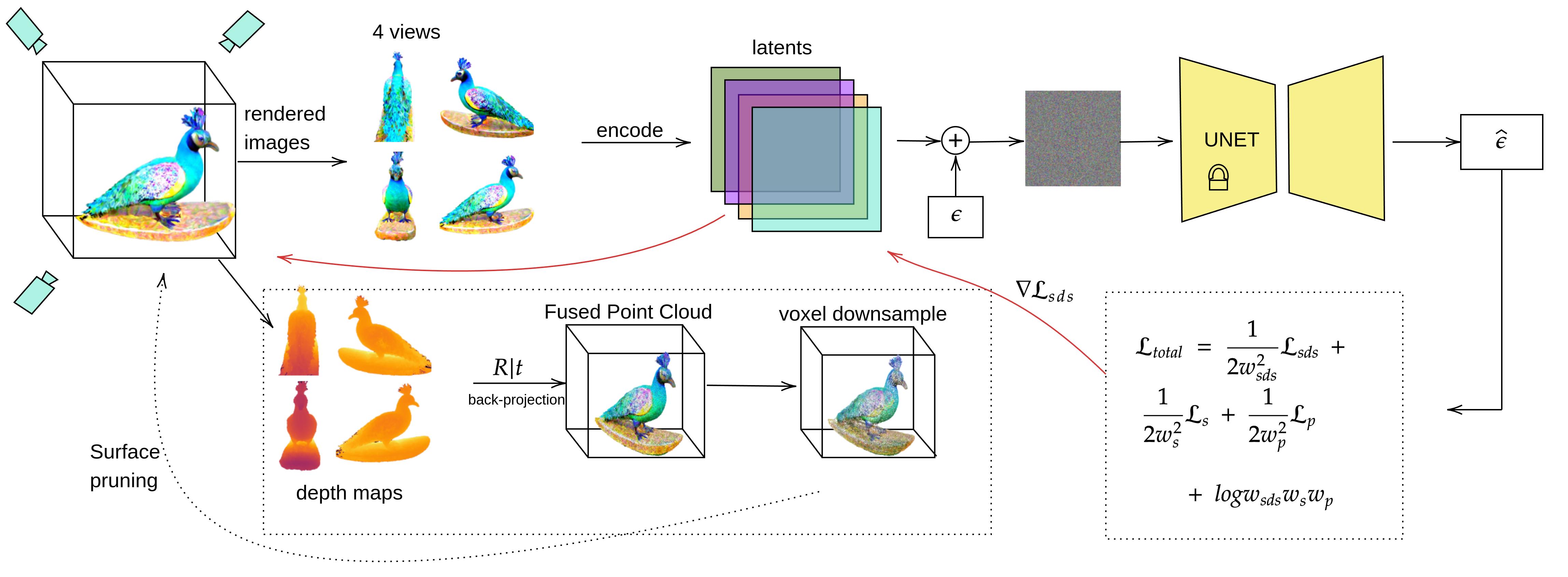}
    \caption{\textbf{Overview of our MVGaussian framework:} Our approach begins with the random initialization of gaussians within a unit sphere, refined iteratively using an SDS-based optimization strategy. gaussians are optimized near the true surface, moving toward the pseudo surface while pruning those farther away. Each iteration renders four views with random azimuth angles, encoded into the latent space. Gaussian noise is added and denoised using a UNET model to compute the loss \(\mathcal{L}_{sds}\). The optimization gradient \(\nabla \mathcal{L}_{sds}\) updates the gaussians, incorporating a feedback loop with fused point cloud data and voxel downsampling to enhance accuracy.}
    \label{fig:overview}
\end{figure*}

\subsection{3D Gaussian Splatting}

The seminal work presented by \citet{3DGS} introduces an explicit approach for representing and rendering three-dimensional objects using gaussian functions as the fundamental building blocks. 
% Unlike traditional mesh-based representations, which rely on discrete vertices and faces,
3D Gaussian splatting employs continuous Gaussian distributions to define the geometry and appearance of a 3D model. Each gaussian is characterized by its center position or mean \( \mu \), color \( c \), opacity \( \sigma \) and a full covariance matrix \( \Sigma \). This covariance matrix \( \Sigma \) is decomposed into a rotation matrix \( R \) and a scaling matrix \( S \), formulated as:
\begin{equation}
    \Sigma = RSS^TR^T    
\end{equation}

The parameters \( R \) and \( S \) are stored and optimized independently. Consequently, a 3D gaussian can be defined as follows:

\begin{equation}
    G(x) = \exp \left( -\frac{1}{2} (x - \mu)^T \Sigma^{-1} (x - \mu) \right)
\end{equation}

where \(x \in \mathbb{R}^3\) represents a point in 3D space, \(\mu \in \mathbb{R}^3\) is the gaussian's center, and \(\Sigma \in \mathbb{R}^{3 \times 3}\) is the covariance matrix that defines the shape, orientation, and scale of the gaussian. Each gaussian splat also includes a color vector \(c \in \mathbb{R}^3\) representing the RGB values and a scalar \(\sigma \in [0, 1]\) indicating the transparency level.

For an arbitrary 3D point \(x\), the influence of a gaussian splat is formulated as:

\begin{equation}
\label{eq:alpha-blending}
    \alpha = \sigma \exp \left( -\frac{1}{2} (x - \mu)^T \Sigma^{-1} (x - \mu) \right)
\end{equation}

To render a scene using gaussian splats, the 3D gaussians are projected onto a 2D image plane. The contribution of each splat to the final image is determined by integrating the gaussian over the pixels it influences.
% The rendered image is computed by blending these contributions based on their opacity.
The final color of a pixel \(p\) is a combination of the influences from all ordered points \( x \) that project to pixel \(p\):

\begin{equation}
    C = \sum_{i} c_i \alpha_i \prod_{j = 1}^{i-1} (1 - \alpha_j)
\end{equation}

Unlike implicit representations used in NeRF models, Gaussian splatting is an explicit method that requires a mechanism to manage the number of gaussians. This is achieved through a unique densification and pruning scheme, discussed subsequently.

%%%%%%%%%%%%%%%%%%%%%%%%%%%%%%%%%%%%%%

\section{Method}

We focus on reducing the Janus problems by successfully generating a new densification and pruning scheme by not only using multi-view guidance but also by backprojecting the points from the estimated depth on the fly to optimize the gaussians. To the best of our knowledge, this is the first approach that aims to optimize the gaussians using the estimated depth. Most current SDS-based techniques only focus on the estimation of the score but do not focus on the gaussians.
Furthermore, we observe the advantages that surface alignment offers in SuGaR \cite{sugar} pertaining to surface generation and mesh extraction. However, unlike SuGaR instead of a post-processing term we introduce a novel \textbf{regularization term} that allows for flattening the gaussians during the learning process itself.
Our approach focuses on the improvement of gaussians by reducing the Janus problem and achieving fast rendering for text-to-3D tasks.% we employ a SDS-based approach using 3DGS representation.  
 We primarily base our method on multi-view guidance to alleviate the issue of Janus and generate consistent 3D results. Further, we remodel the densification strategy and enforce surface-closeness for the generated gaussian, drawing inspiration from prior methods. %Further, we ensure flatness and high opacity through a regularization term. 

The overall framework, depicted in Figure \ref{fig:overview}, showcases how our approach integrates these components to produce more consistent and unified 3D reconstructions. This method not only addresses the shortcomings of prior techniques but also enhances the efficiency and quality of the 3D models generated, offering a viable solution to the Janus problem in SDS-based methods.

\begin{figure*}[!htp]
    \centering
    \setlength{\tabcolsep}{-1pt}
    \scalebox{0.9}{
    \begin{tabular}{cccccccc}
        % \textbf{Original} & \textbf{Ours} & \textbf{Original} & \textbf{Ours} \\
        GaussianDreamer & GSGen & LucidDreamer & \textbf{MVGaussian}  & 
        GaussianDreamer & GSGEN & LucidDreamer & \textbf{MVGaussian} \\
        &&&(ours)&&&&(ours)\\
        % (MVDream Guidance) & & & & (MVDream Guidance) & & &\\ 
        \multicolumn{4}{c}{``An armored green-skin orc warrior } & \multicolumn{4}{c}{
        ``A forbidden castle high up in the mountains" 
         } \\
        \multicolumn{4}{c}{riding a vicious hog"} & \\
        \includegraphics[width=0.15\linewidth]{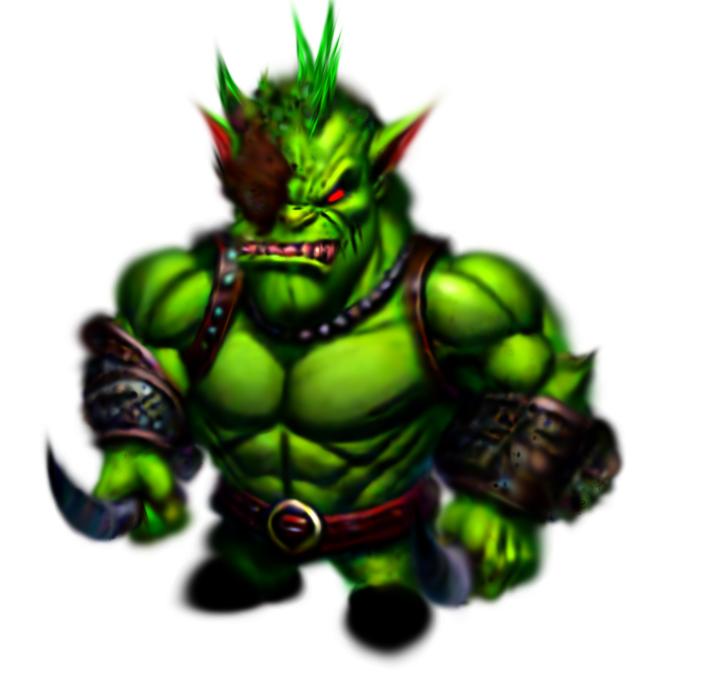} &

        \includegraphics[width=0.15\linewidth]{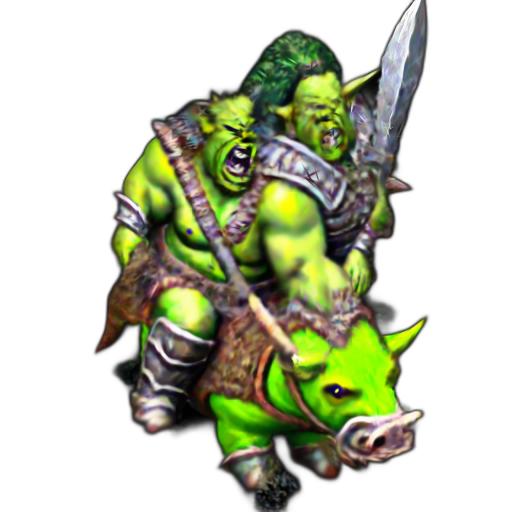} 
        
        &

        \includegraphics[width=0.1\linewidth]{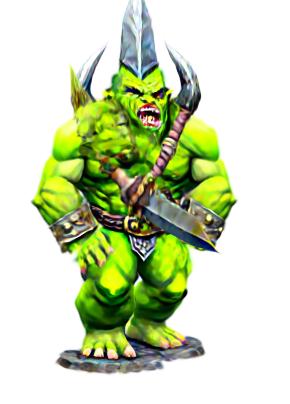}

        &

        \includegraphics[width=0.13\linewidth]{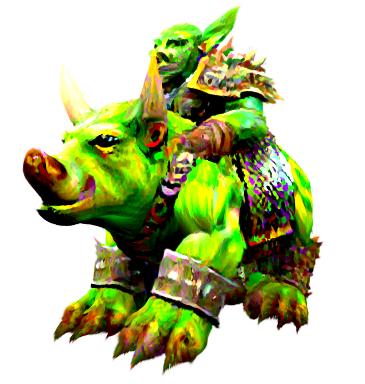} &

        \includegraphics[width=0.13\linewidth]{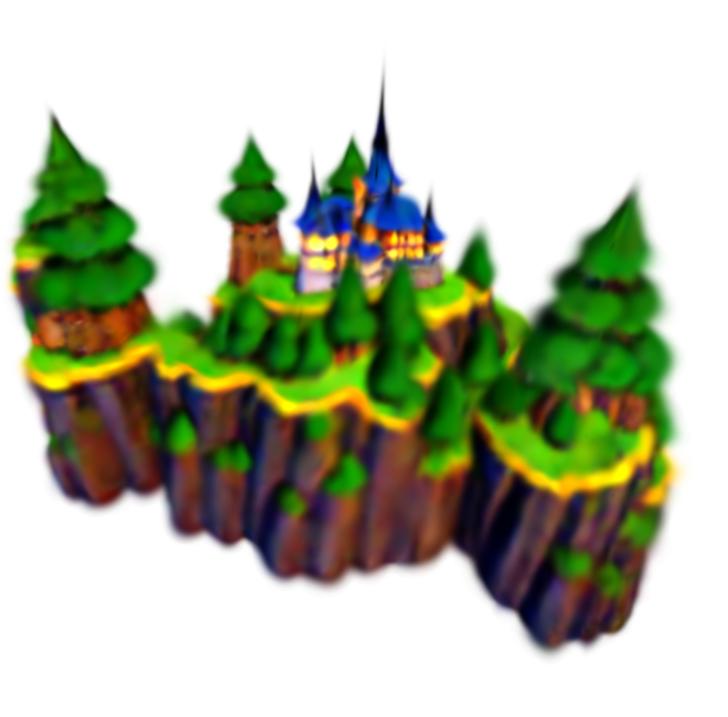} 
        &
        
        \includegraphics[width=0.12\linewidth]{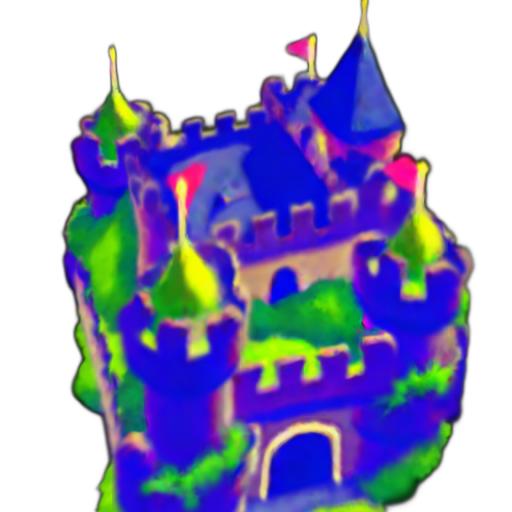}   
        &  
        \includegraphics[width=0.14\linewidth]{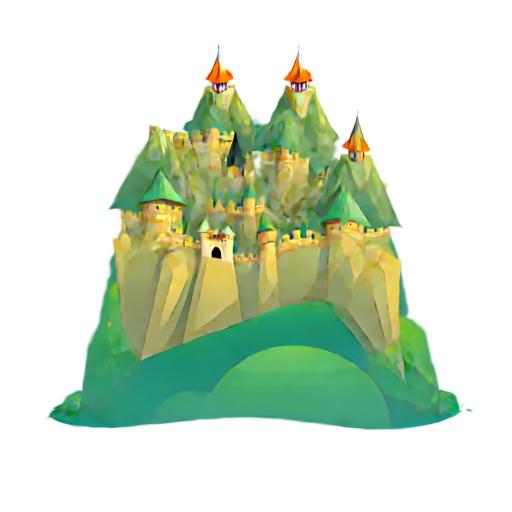}   
        &
        \includegraphics[width=0.15\linewidth]{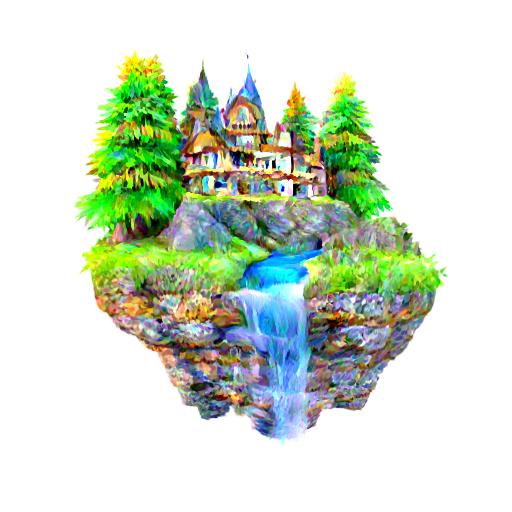}

        \\
        \multicolumn{4}{c}{"A flying dragon, highly detailed, realistic, majestic."} &  \multicolumn{4}{c}{"A 3D model of an adorable cottage } \\

        \multicolumn{4}{c}{}& \multicolumn{4}{c}{with a thatched roof"} \\
        % \includegraphics[width=0.1\linewidth]{gscraft_no_regularization/paris/gaussian_dreamer/41_cropped.jpg} &  
        % \includegraphics[width=0.12\linewidth]{gscraft_no_regularization/paris/gsgen/rgb_15000_0023_rgba.jpg}   &

        % \includegraphics[width=0.12\linewidth]{gscraft_no_regularization/paris/lucid_dreamer/render_view_8.jpg}
        
        % &
        % \includegraphics[width=0.14\linewidth]{gscraft_no_regularization/paris/ours/15_frame.jpg}   
        % &

        \includegraphics[width=0.14\linewidth]{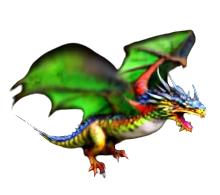} &  
        \includegraphics[width=0.14\linewidth]{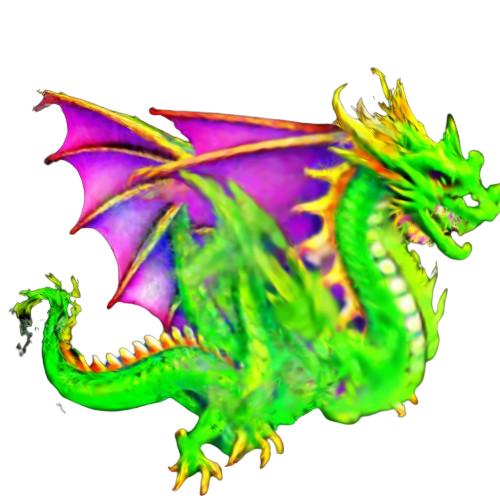} &
        \includegraphics[width=0.14\linewidth]{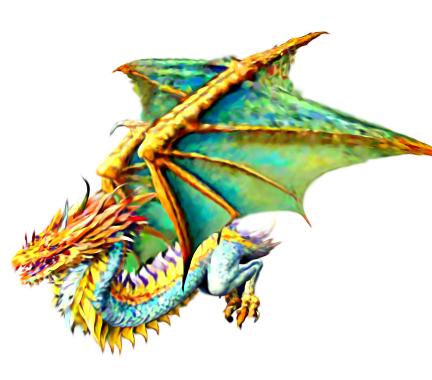} &
        \includegraphics[width=0.14\linewidth]{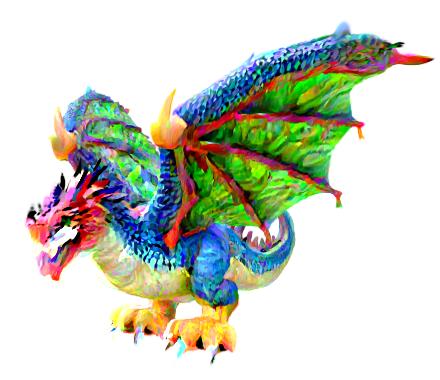} &  
        
        \includegraphics[width=0.14\linewidth]{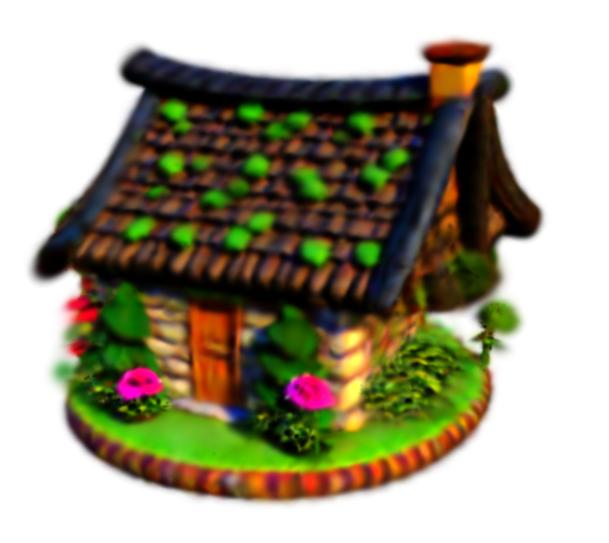} &  
        \includegraphics[width=0.12\linewidth]{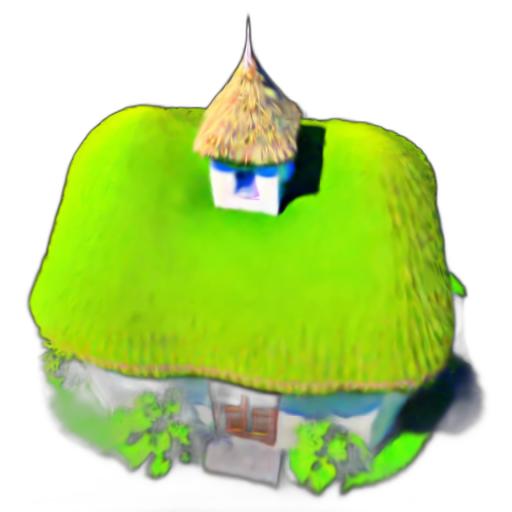}  &  

        \includegraphics[width=0.13\linewidth]{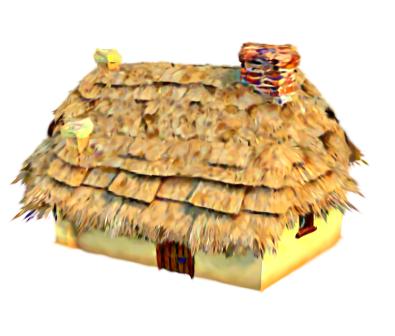}
        
        &
        \includegraphics[width=0.14\linewidth]{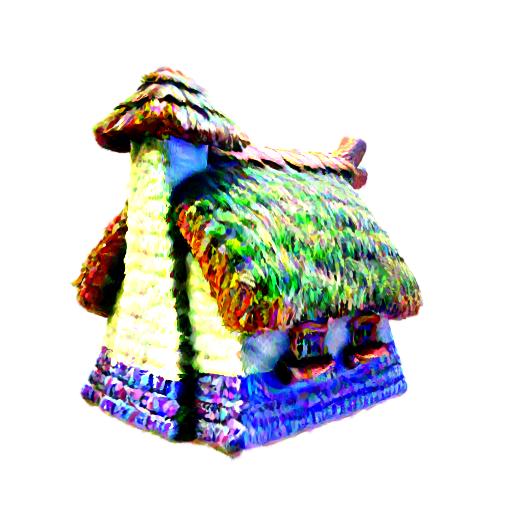}

        \\

        \multicolumn{4}{c}{"A blue jay sitting on a willow basket of macarons"} & \multicolumn{4}{c}{"Medieval soldier with shield and sword, fantasy, game, character, } \\
        \multicolumn{4}{c}{} & \multicolumn{4}{c}{ highly detailed, photorealistic, 4K, HD"}\\
        \includegraphics[width=0.1\linewidth]{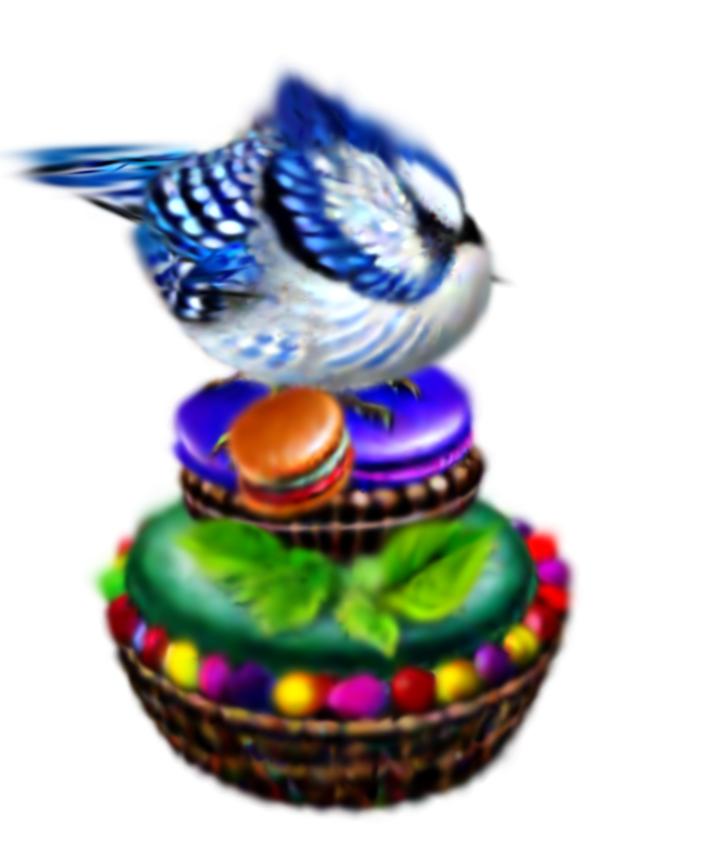} & 
        
        \includegraphics[width=0.1\linewidth]{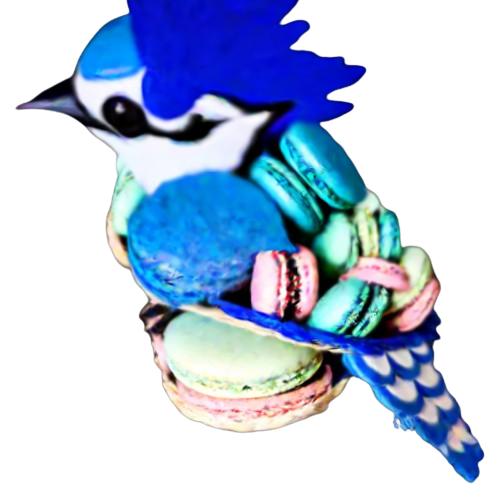} & 

        \includegraphics[width=0.10\linewidth]{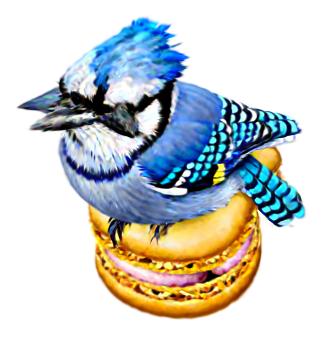} &
        
        \includegraphics[width=0.13\linewidth]{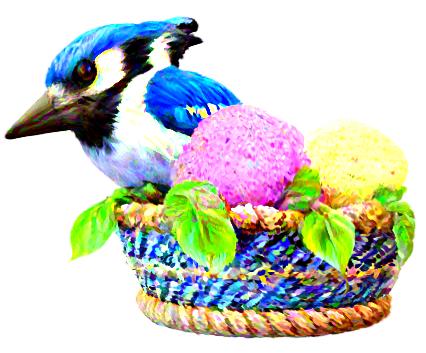} &
    
        \includegraphics[width=0.14\linewidth]{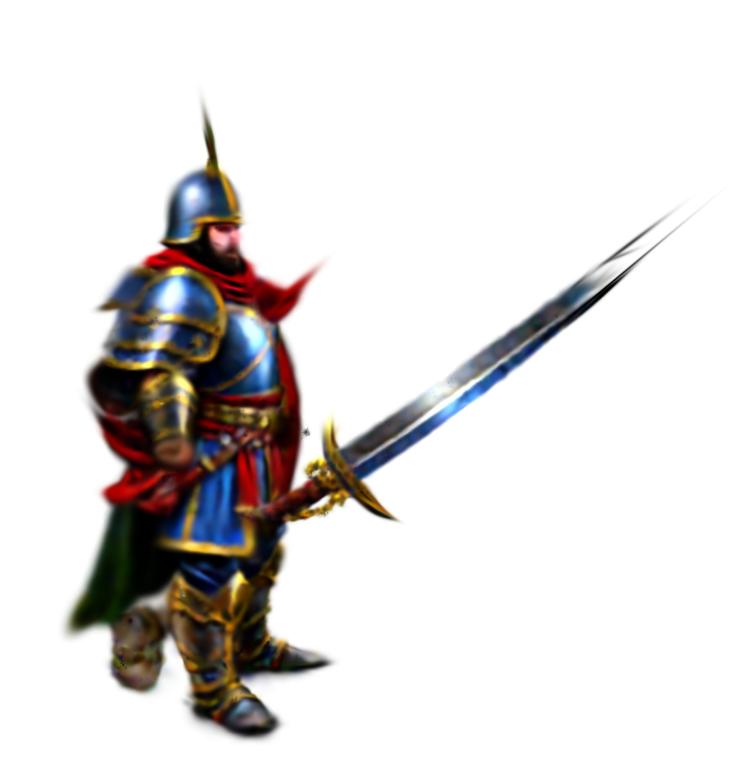} &

        \includegraphics[width=0.12\linewidth]{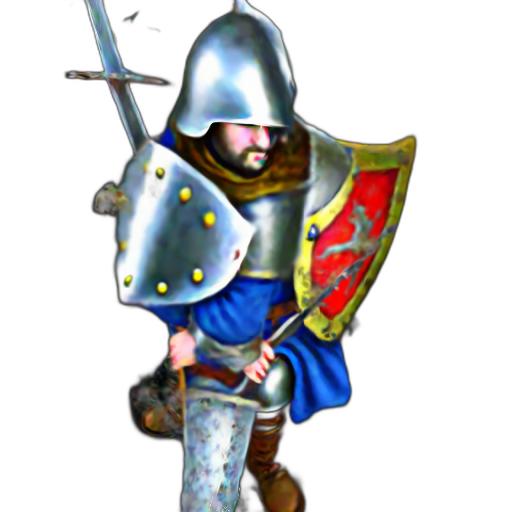}   
        & 

        \includegraphics[width=0.1\linewidth]{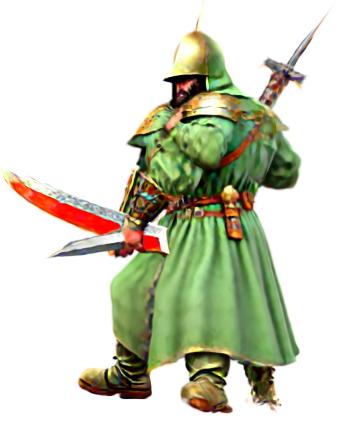}   
        
        &
        \includegraphics[width=0.135\linewidth]{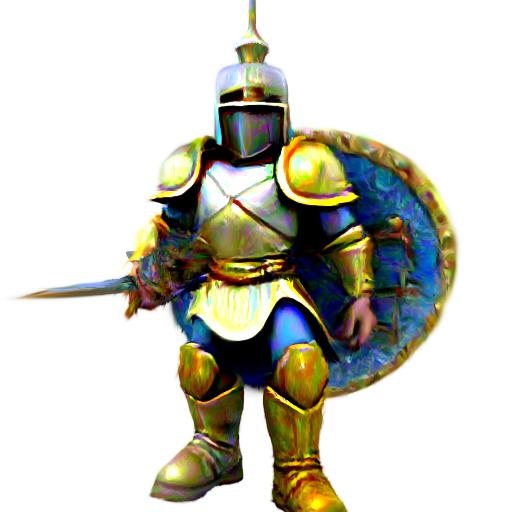}   
        \\

        \multicolumn{4}{c}{"Jack Sparrow wearing sunglasses, head, photorealistic, 8K, HD, raw" } & \multicolumn{4}{c}{"A peacock standing on a surfing board, } \\
        \multicolumn{4}{c}{} & \multicolumn{4}{c}{ highly detailed, majestic"} \\
        
        \includegraphics[width=0.13\linewidth]{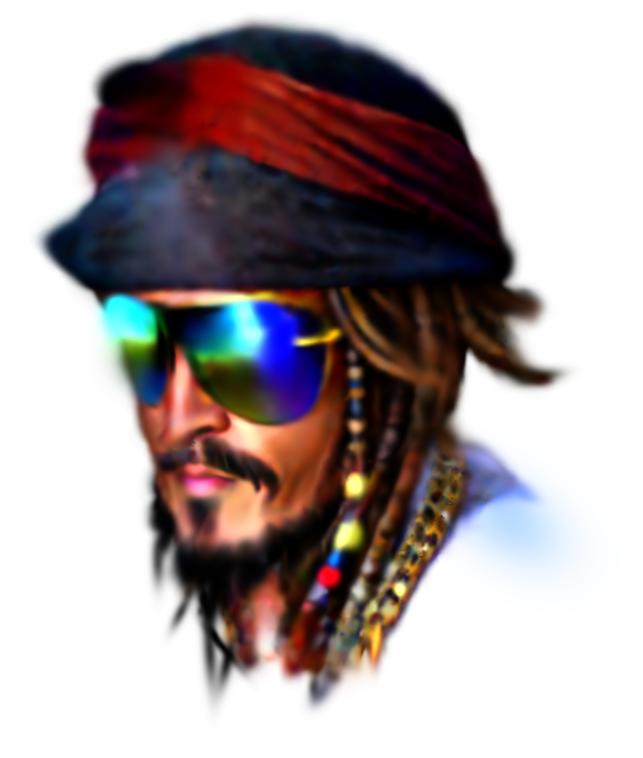} &  

        \includegraphics[width=0.14\linewidth]{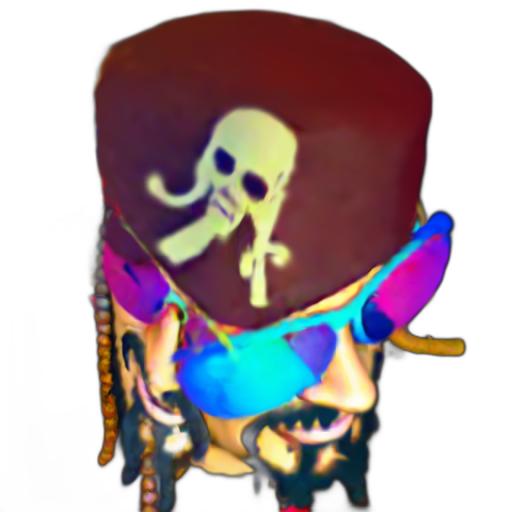}
        & 
        \includegraphics[width=0.14\linewidth]{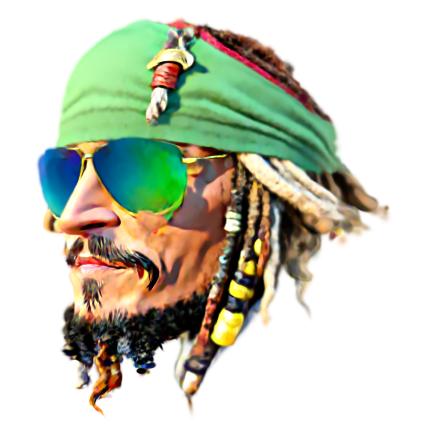}
        &

        \includegraphics[width=0.14\linewidth]{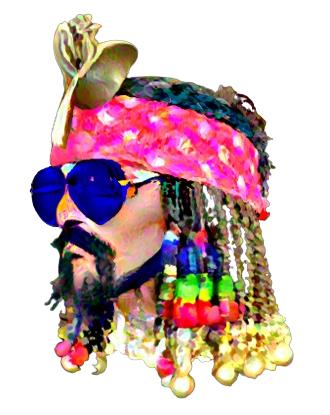}   &

        \includegraphics[width=0.14\linewidth]{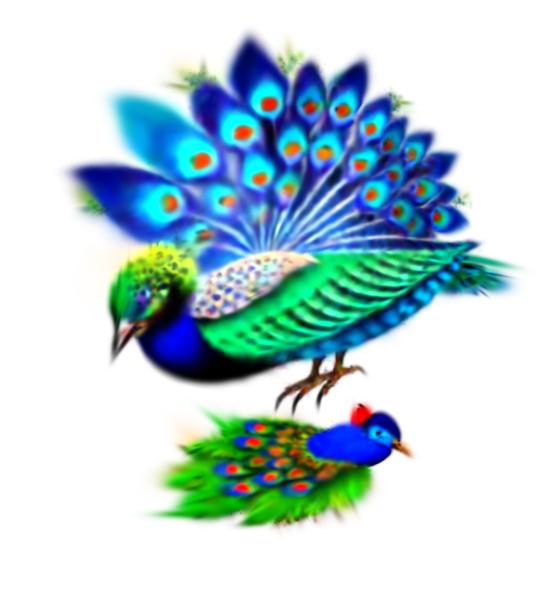}
         
         &  
        \includegraphics[width=0.12\linewidth]{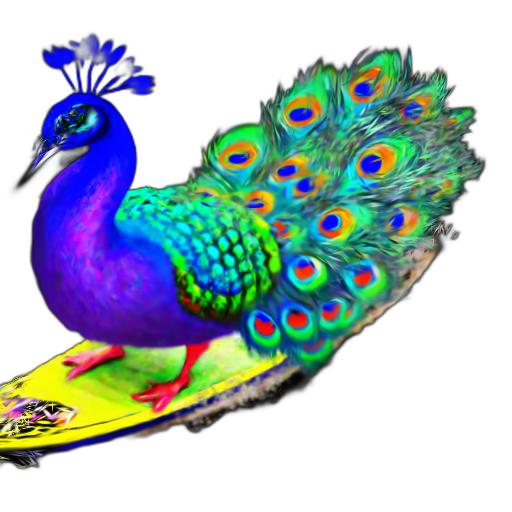}
        & 
        \includegraphics[width=0.16\linewidth]{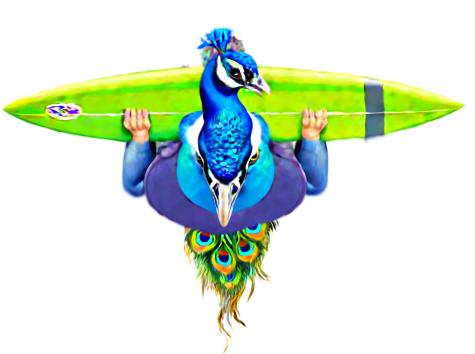}
        &
        \includegraphics[width=0.16\linewidth]{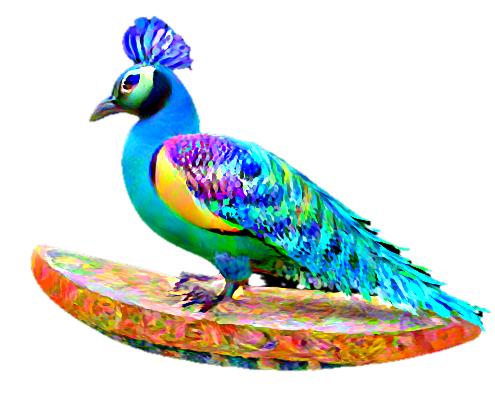}  
    \end{tabular}}
    \caption{We show extensive qualitative results in the figure above and show comparisons against several state-of-the-art methods. We show consistent improvement across all different prompts tested and demonstrate the effectiveness of our densification approach.}
    \label{fig:comparisons}
\end{figure*}

\begin{figure*}[!htp]
    \setlength{\tabcolsep}{1pt}
    \scalebox{0.9}{
    \begin{tabular}{c c c c c } 
         \textbf{Ours}  & Fantasia3D & Prolific Dreamer & LucidDreamer & Magic3D  
         \\
        ($\sim 25mins$) & ($\sim 1hr$) & ($\sim 8hrs$) & ($\sim 35mins$) & ($\sim 1hr$)\\
         \multicolumn{5}{c}{“A DSLR photo of the Imperial State Crown of England.”}\\
         \includegraphics[width=0.15\linewidth]{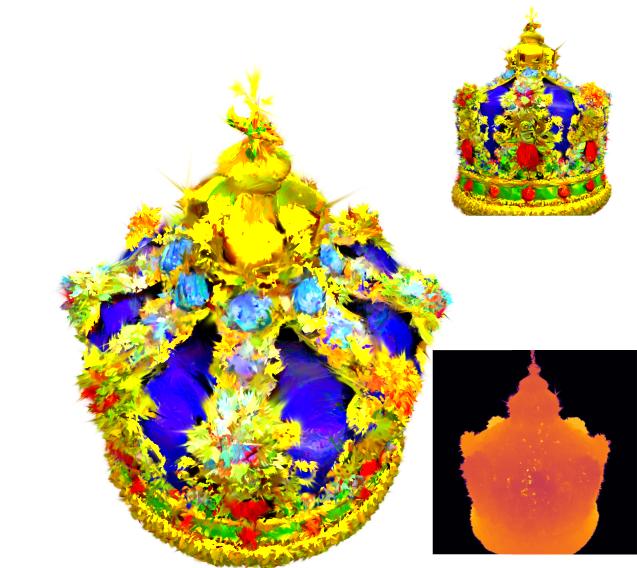} & 
         \includegraphics[width=0.18\linewidth]{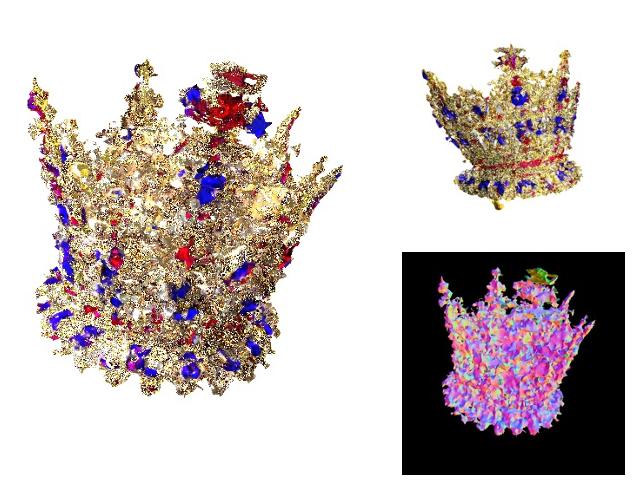} &
         \includegraphics[width=0.16\linewidth]{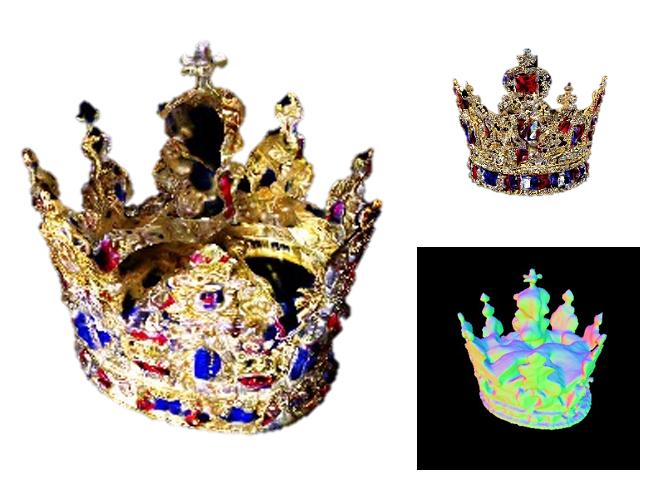} & 
         \includegraphics[width=0.18\linewidth]{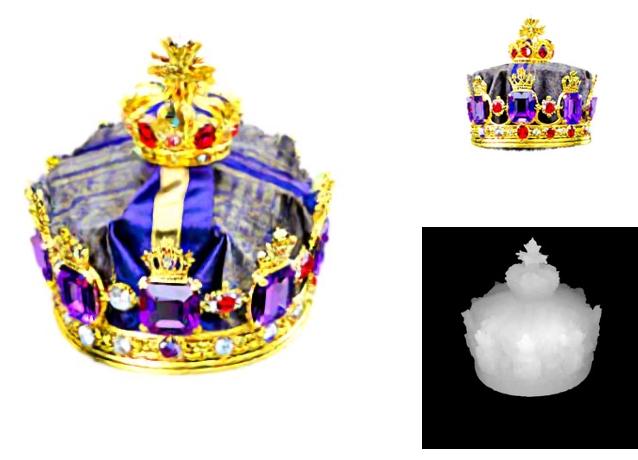} & 
         \includegraphics[width=0.2\linewidth]{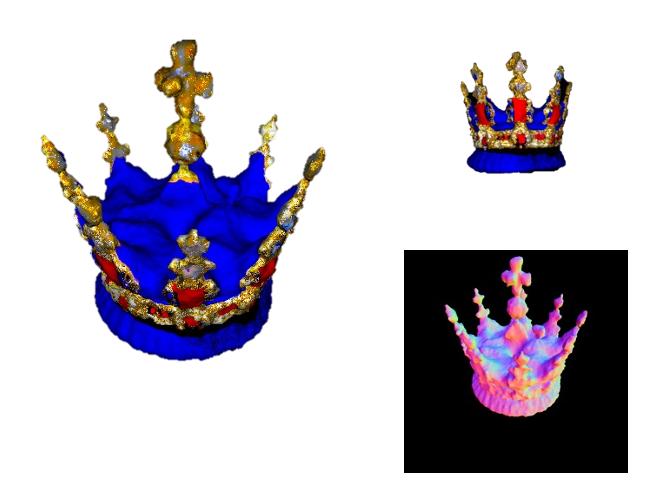}
         \\

        \multicolumn{5}{c}{“A DSLR photo of a Schnauzer wearing a pirate hat.”}\\
         \includegraphics[width=0.15\linewidth]{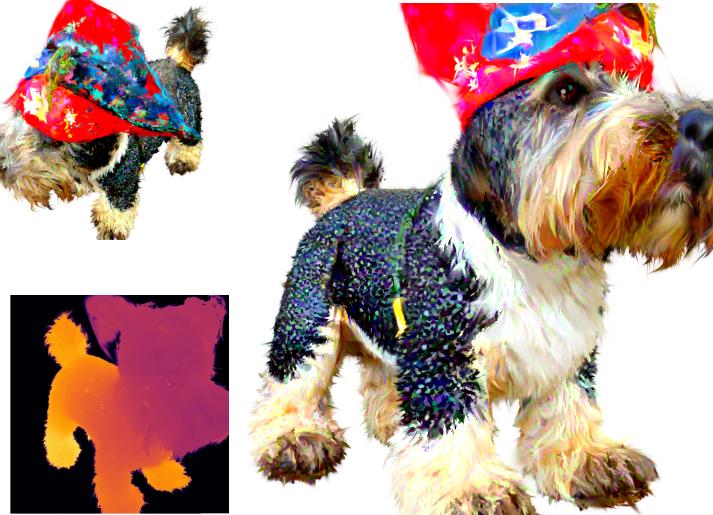} & 
         \includegraphics[width=0.2\linewidth]{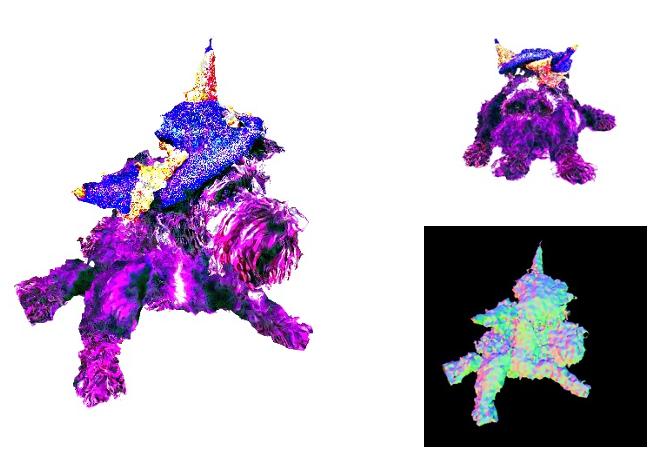} &
         \includegraphics[width=0.2\linewidth]{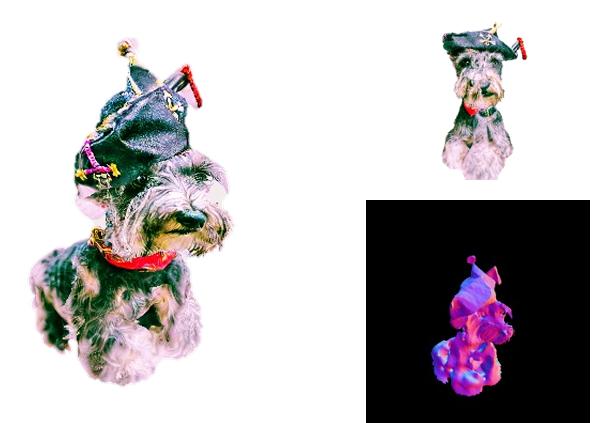} & 
         \includegraphics[width=0.2\linewidth]{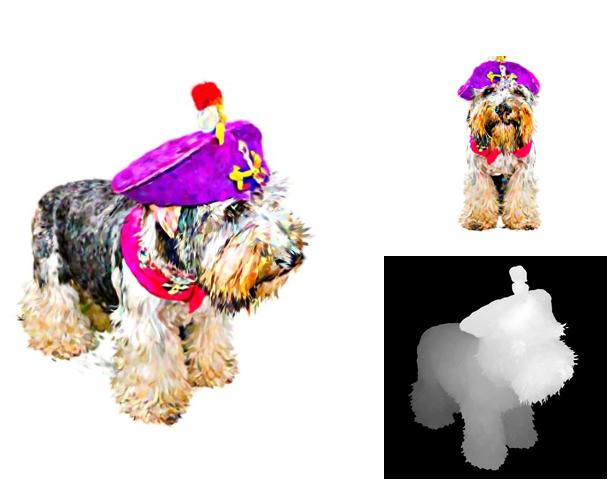} & 
         \includegraphics[width=0.2\linewidth]{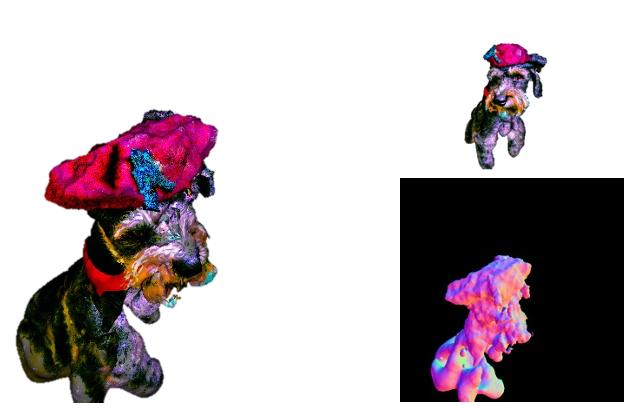}
         
    \end{tabular}}
    \caption{Additional qaulitative comparisons with several state-of-the-art methods.}
    \label{fig:comparison_fant}
\end{figure*}

\subsection{Multi-view guidance for consistent 3D generation}

% TODO: introduce Janus problem earlier?
SDS-based approaches for text-to-3D generation often suffer from multi-face or the Janus problem. This issue arises as diffusion models are trained on 2D images and lack a true understanding of the 3D world. Consequently, while rendered images might appear plausible from different viewpoints, they often fail to represent a consistent and unified 3D object. Several strategies have been developed to address the Janus problem. Notably, Zero123 \cite{liu2023zero} and MVDream \cite{mvdream} have made significant strides by fine-tuning pre-trained diffusion models on 3D data. Zero123 predicts multi-view images conditioned on a reference image and camera position, while MVDream fine-tunes diffusion models to generate multi-view images from text inputs. Despite these advancements, these methods do not completely resolve the Janus problem, as the generated multi-view images often lack the exact consistency needed for unified 3D models since they lack precise symmetry and high-level match of details across the generated views. However, they provide reliable guidance for SDS-based approaches.

To address the Janus problem, we integrate the strengths of MVDream as the primary guidance mechanism within our framework. By adopting MVDream, we leverage its ability to generate multi-view images from textual inputs, thereby providing robust guidance for our 3D models. As shown in Figure \ref{fig:overview}, we render 4 views around the current object at each training step. We then map these muli-view images to the latent space and perform the noising and denoising steps. Similar to Dreamfusion, we adopt classifier-free guidance (CFG) proposed by \citet{CFG} to enhance the quality of generated 3D models. CFG adjusts the score function to favor regions with a higher ratio of conditional to unconditional density, using a guidance scale parameter \(\omega\). 
\begin{comment}
    
%%%%% this equation is well known can be skipped I guess %%%%%
This modifies the noise prediction as follows:
\begin{align*}
    \hat{\epsilon}_\phi(z_t; y, t) = (1 + \omega) \epsilon_\phi(z_t; y, t) - \omega \epsilon_\phi(z_t; t)
\end{align*}
in which \( y \) is the text embeddings, \( t \) is the time step in the diffusion process, \(\hat{\epsilon}\) and \(\hat{p}\) represent the guided versions of noise prediction and marginal distribution, respectively. Setting \(\omega > 0\) improves the sample fidelity. This predicted noise $ \hat{\epsilon} $ is then used to compute the $\mathcal{L}_{sds}$ loss.
\end{comment}
% Our framework combines the explicit representation advantages of 3D gaussian Splatting with the enhanced multi-view consistency offered by MVDream. This combination not only mitigates the Janus problem but also enhances the overall quality and fidelity of the 3D reconstructions. Additionally, our approach demonstrates improved performance in capturing fine details and intricate geometries, further advancing the state of the art in text-to-3D generation.
Further, we look into gaussian alignment to improve the rendering.
\subsection{Gaussian Alignment for Optimal Geometry}
Inspired by \citet{sugar}, we propose a novel, simpler regularization term that is faster and easier to optimize. The authors demonstrate the effectiveness of aligning gaussians to the surface using a regularization term \(\mathcal{R}\) that minimizes the Signed Distance Function (SDF) of the gaussians to the surface. However, this method may not be suitable for score distillation strategies, as it requires prior SDF estimation before appearance modeling, as shown in Fantasia3D \cite{fantasia3d}. Additionally, the SuGaR method does not optimize the gaussians on the fly but acts as a post-processing approach.

Suppose we have a true surface of the 3D scene described by a given text prompt; we want the gaussians to lie on the surface to capture fine details and intricate geometries, resulting in high-fidelity reconstructions. For any point \( x \in \mathbb{R}^3 \) on the surface, based on Eq. \ref{eq:alpha-blending}, we can find the gaussian \( g^* \) that has the most significant influence on the appearance of \( x \):

\begin{align}    
 g^* = &\arg\max_g \left[ \sigma_g \exp \left( -\frac{1}{2} (x - \mu_g)^T \Sigma_g^{-1} (x - \mu_g) \right) \right] \>
 \label{eq:g_star}
\end{align}

Ideally, we want the center of \( g^* \) to be close to the surface point \( x \), i.e., \( \mu_g \rightarrow x \), which makes the exponent approach 0:

\begin{align}    
    T = (x - \mu_{g^*})^T \Sigma^{-1}_{g^*} (x - \mu_{g^*}) \rightarrow 0
    \label{eq:exponent}
\end{align}
% \textit{
Thus, minimizing this exponent term encourages the gaussians to be close to the surface. When a gaussian lies on a surface, we want it to be flat to assemble the geometry of the 3D object correctly. Henceforth, we drive one of the three scales of the gaussian \( g^* \) should be close to 0. We can express \( \Sigma_{g^*} \) in terms of its eigenvalues and eigenvectors:
\[
\Sigma_{g^*} = U \Lambda U^T
\]

where \( U \) is the matrix of eigenvectors and \( \Lambda \) is the diagonal matrix of eigenvalues.
% \begin{comment}

\begin{equation}
    U = [\mathbf{v}_1 \; \mathbf{v}_2 \; \mathbf{v}_3]; \quad \Lambda = \begin{pmatrix}
    \lambda_1 & 0 & 0 \\
    0 & \lambda_2 & 0 \\
    0 & 0 & \lambda_3
    \end{pmatrix}
\end{equation}

The inverse of the covariance matrix \( \Sigma_g \) is given by:

\begin{equation}
    \Sigma_{g^*}^{-1} = U \Lambda^{-1} U^T = \begin{pmatrix} \mathbf{v}_1 & \mathbf{v}_2 & \mathbf{v}_3 \end{pmatrix} \begin{pmatrix} \frac{1}{\lambda_1} & 0 & 0 \\ 0 & \frac{1}{\lambda_2} & 0 \\ 0 & 0 & \frac{1}{\lambda_3} \end{pmatrix} \begin{pmatrix} \mathbf{v}_1^T \\ \mathbf{v}_2^T \\ \mathbf{v}_3^T \end{pmatrix}
\end{equation}
% \end{comment}
Substituting into the exponent term $T$ in Eq. (\ref{eq:exponent}), we have:

\begin{equation}
    T = \sum_{i} \frac{1}{\lambda_i} (x - \mu_g)^T \mathbf{v}_i \mathbf{v}_i^T (x - \mu_g)
\end{equation}
where $v_i$ are the eigenvectors in $U$.

If the scale in direction \(i \in [1,2,3]\) of a gaussian is close to 0, its corresponding eigenvalue \(\lambda_i\) is also very small. Thus, the exponent is governed and can be approximated by the term:
% \( \frac{1}{\lambda_i} (x - \mu_g)^T \mathbf{v}_i \mathbf{v}_i^T (x - \mu_g) \):

\begin{equation}
    T \approx \frac{1}{\lambda_i} (x - \mu_g)^T \mathbf{v}_i \mathbf{v}_i^T (x - \mu_g)
\end{equation}

Hence, to encourage the gaussians to be flat on the surface, we minimize the following loss term:

\begin{equation}
    \mathcal{L}_{s} = \sum_g \frac{1}{\lambda_g} (x - \mu_g)^T \mathbf{v}_g \mathbf{v}_g^T (x - \mu_g)
\end{equation}

where \(\lambda_g\) and \(\mathbf{v}_g\) are the eigenvalue and eigenvector corresponding to the smallest scale of the gaussian \(g\).

To further ensure that the Gaussians align closely with the surface, we introduce a surface proximity regularization term. This term minimizes the squared distance between the Gaussian centers \(\mu_g\) and the nearest points on the true surface \(p_i\). The loss is defined as:

\[
\mathcal{L}_{\text{p}} = \sum_g \sum_i \alpha_{gi} \| p_i - \mu_g \|^2
\]

where \(\alpha_{gi}\) are weights that ensure each Gaussian center \(\mu_g\) is associated with the nearest surface point \(p_i\). These weights are computed using a soft assignment based on the distance:

\[
\alpha_{gi} = \frac{\exp(-\| p_i - \mu_g \|^2 / \tau)}{\sum_j \exp(-\| p_j - \mu_g \|^2 / \tau)}
\]

where \(\tau\) is a temperature parameter that controls the sharpness of the assignment. By minimizing this term, we encourage the Gaussians to be positioned close to the surface, further enhancing the alignment of the Gaussians with the true geometry of the 3D object. Since computing this term requires calculating distances between Gaussian centers and surface points, to make this computationally feasible, a subset of sampled points from both Gaussian centers and surface points can be used, reducing the computational burden while still maintaining the effectiveness of the regularization.

\begin{comment}
Another factor that influences a gaussian's impact on a point is its opacity. Thus, we encourage each gaussian to have high opacity by adding another loss term:

\begin{equation}
    \mathcal{L}_\sigma = \sum_g \| 1 - \sigma_g \|^2
\end{equation}

where \(\sigma_g\) is the opacity of the gaussian \(g\).
\end{comment}
The final loss is the weighted sum of the individual losses:

\begin{equation}
    \mathcal{L}_{total} = w_{sds} \mathcal{L}_{sds} + w_{s} \mathcal{L}_{s} + w_{p} \mathcal{L}_{p}
\end{equation}
which contains a factor that influences a gaussian's opacity. 
To effectively balance the contributions of our three loss terms, \(\mathcal{L}_{sds}\), \(\mathcal{L}_{s}\), and \(\mathcal{L}_{o}\), we employ a homoscedastic uncertainty-based weighting approach \cite{UncertaintyLoss}. The total loss, \(\mathcal{L}_{total}\), is formulated by introducing learnable weights \(w_{sds}\), \(w_{s}\), and \(w_{o}\) corresponding to each loss term. These weights dynamically scale the losses according to their estimated uncertainty, as given by:

\[ \mathcal{L}_{total} = \frac{1}{2w_{sds}^2} \mathcal{L}_{sds} + \frac{1}{2w_{s}^2} \mathcal{L}_{s} + \frac{1}{2w_{p}^2} \mathcal{L}_{p} + \log w_{sds}w_{s}w_{p} \]

By optimizing the weights \(w_{sds}\), \(w_{s}\), and \(w_{o}\) during training, the model balances the magnitude of the individual loss terms, leading to more stable and effective learning.

\subsection{Surface densification and pruning}
% We use MVDream guidance for training the gaussian splat representation. 
In this section, we further relook at the densification strategy used in 3DGS and discuss our strategy to overcome the limitations of the current methods.
Naive 3D Gaussian splatting methods densify the gaussians based on the gradient of the gaussian centers and the scales of the gaussians. While this approach is straightforward, it presents several significant drawbacks. One of the primary challenges lies in defining an appropriate threshold value for the gradient. If the threshold value is set too high, fewer gaussians are added to the scene, leading to a lack of detail in the reconstructed model. Conversely, if the threshold value is set too low, the number of gaussians increases significantly. This not only hinders the learning speed but also impedes the convergence of the model due to the excessive computational load.

We propose an intuitive method that utilizes the rendered image and depth to backproject the rendered pixels to the world using camera parameters. This allows us to progressively reconstruct the surface of the 3D model.

% Based on this reconstructed surface, we prune gaussians that are too far from the surface, as their contribution to the appearance is likely diminished by overlapping gaussians. 
We densify the gaussians that are close to the surface, allowing the model to gradually reconstruct the missing parts and speed up the training time due to the significantly reduced number of gaussians to update.

Mathematically, we define the backprojection of a pixel \( p \) with depth \( d \) and camera parameters \( K \) (intrinsic matrix) and \( [R|t] \) (extrinsic matrix) as follows:

\[
P = R^{-1}(K^{-1} p' d - t)
\]

where \( p' \) is the homogeneous coordinate of \( p \). Let \( \{ P_i \} \) be the set of all backprojected points. We then define the distance \( D_g \) of a gaussian \( g \) from the surface as the Euclidean distance between the gaussian center \( \mu_g \) and the closest backprojected point \( P_* \):

\[
D_g = \min_{P_i} \| \mu_g - P_i \|
\]

We prune gaussians for which \( D_g \) exceeds a threshold \( \epsilon \).
This approach allows us to significantly reduce the number of gaussians and improve the efficiency and quality of the final 3D reconstruction.

\begin{table*}[ht]
\centering
\scalebox{0.7}{
\begin{tabular}{|p{10cm}|c|c|c|c|c|c|}
\hline
\textbf{Prompt} & \multicolumn{4}{c|}{\textbf{Human evaluation scores}} & \multicolumn{2}{c|}{\textbf{No. Gaussians (million)}} \\ \cline{2-7} 
 & \textbf{GaussianDreamer} & \textbf{Gsgen} & \textbf{Lucid Dreamer} & \textbf{Ours} & \textbf{Naive} & \textbf{Ours} \\ \hline
``A blue jay sitting on a willow basket of macarons" & 3.23 & 2.89 & 3.07 & \textbf{4.65} & 16.2 & \textbf{1.1} \\ \hline
``A flying dragon, highly detailed, realistic, majestic." & 2.51 & 2.12 & 3.45 & \textbf{4.81} & 24.5 & \textbf{1.2} \\ \hline
``An armored green-skin orc warrior riding a vicious hog." & 3.12 & 2.94 & 3.21 & \textbf{4.56} & 22.7 & \textbf{1.3} \\ \hline
``A forbidden castle high up in the mountains." & 3.24 & 2.46 & 2.87 & \textbf{4.45} & 24.2 & \textbf{1.2} \\ \hline
``A peacock standing on a surfing board, highly detailed, majestic." & 2.02 & 3.56 & 2.95 & \textbf{4.64} & 18.8 & \textbf{1.1} \\ \hline
``Jack Sparrow wearing sunglasses, head, photorealistic, 8k, HD, raw." & 4.01 & 3.21 & 4.15 & \textbf{4.45} & 16.7 & \textbf{0.9} \\ \hline
``Medieval soldier with shield and sword, fantasy, game, character, highly detailed, photorealistic, 4K, HD" & 3.5 & 2.57 & 3.64 & \textbf{4.32} & 17.3 & \textbf{1.2} \\ \hline
``A 3D model of an adorable cottage with a thatched roof" & 3.45 & 1.89 & 3.42 & \textbf{3.89} & 16.9 & \textbf{1.2} \\ \hline
\end{tabular}
}
\caption{Comparison of different methods based on the provided prompts. The table includes human evaluation scores for each method, along with the number of Gaussians (in millions) utilized by both the naive and our approach. The highest scores in each row are highlighted in bold.}
\label{tab:comparison_human_eval}
\end{table*}

\section{Experiments and Results}

We generate a variety of outputs using a diverse set of prompts and observe that our method significantly outperforms other 3DGS-based methods within the same time constraints. Our approach not only achieves a higher level of detail but also exhibits fewer artifacts compared to existing methods. Moreover, by utilizing guidance from MVDream, our method effectively reduces the Janus problem. As shown in Figure \ref{fig:comparisons}, our method produces brighter colors and sharper structures, achieving a photorealistic appearance. Additional results are presented in Figure \ref{fig:comparison_fant}. Our training process involves 10,000 steps, with the densification process starting at 1,000 steps and performed every 200 iterations on an NVIDIA A100. The code is written using PyTorch. The entire process takes approximately 25 minutes for a given prompt.

We compare our method against several state-of-the-art techniques, including GSGen \cite{gsgen}, LucidDreamer \cite{luciddreamer}, Fantasia 3D \cite{fantasia3d}, Magic3D \cite{lin2023magic3d}, and ProlificDreamer \cite{prolificdreamer}, as illustrated in Figure \ref{fig:comparisons} and Figure \ref{fig:comparison_fant}. Our findings indicate that methods such as GSGen and LucidDreamer, which rely heavily on Point-e, struggle to produce high-quality results if the initial point cloud generated by Point-e is suboptimal. For instance, GSGen still exhibits the Janus problem, particularly evident in the \textit{Jack Sparrow} model, while LucidDreamer produces extraneous arms holding the surfboard in the \textit{Peacock} model.

In contrast, our method excels in generating photorealistic results with a higher level of detail compared to other approaches. We observe that Point-e initialization often leads to missing structures, such as the absence of the hog in the \textit{Green Orc} model and the lack of mountains in the \textit{Castle} model generated by GSGen. As shown in Figure \ref{fig:comparisons}, our method consistently uses fewer gaussians $\sim$ 1M to render higher-quality details while other methods require approximately $15-20\times$ more gaussians to achieve the same as shown in Table \ref{fig:comparison_fant}. Furthermore, in most cases, our method achieves a higher CLIP score. For the ones that achieve a lower score, we discuss further in the next section.

\section{Limitations} \label{sec:limitations}
While our method generates high-quality results in a relatively short time, it has several limitations. 
% . \ref{fig:2dvs3d}
One limitation that we observe is the presence of slight over-colorization as can be observed from \textit{Peacock} and \textit{Dragon} models in Figure \ref{fig:comparisons}. The colors generated can be over-saturated and can lack diversity. For a given prompt we do not observe much variation in the objects generated due to the guidance scheme used. Further, methods such as VSD \cite{prolificdreamer} could be used to alleviate this issue.
% Despite achieving the best CLIP score for both 0.33 and 0.32 respectively, while others can only achieve a maximum CLIP score of 0.3 and 0.29 
For several of these models, we also observe small spike-like artifacts such as in the \textit{Dragon} model.

% Another limitation is the evaluation criteria. Despite notable visual improvements, the CLIP score does not accurately reflect the quality of the images. For example, in Figure \ref{fig:comparisons}, the \textit{Castle} appears more photorealistic and highly detailed, yet it receives a lower CLIP score of 0.3 as compared to a score of 0.32 achieved by GaussianDreamer.
% Similarly, for the \textit{Green Orc}, GSGen, despite having the Janus problem, scores the same as our method with a score of 0.31.
We find the CLIP score to be unreliable as further discussed in the appendix in Table  \ref{tab:comparison} as an evaluation criterion and thus conduct a user study to further establish the effectiveness of our method.

\begin{comment}
\section{Limitations} \label{sec:limitations}
While our method generates high-quality results in a relatively short time, it has several limitations. 
% . \ref{fig:2dvs3d}
One limitation that we observe is the presence of slight over-colorization at times as can be observed from \textit{Peacock} and \textit{Dragon} models in Figure \ref{fig:comparisons}. Despite achieving the best CLIP score for both 0.33 and 0.32 respectively, while others can only achieve a maximum CLIP score of 0.3 and 0.29 for these models we observe small spike-like artifacts in the \textit{Dragon} model further discussed in the supplementary.
Another limitation is the evaluation criteria. Despite notable visual improvements, the CLIP score does not accurately reflect the quality of the images. For example, in Figure \ref{fig:comparisons}, the \textit{Castle} appears more photorealistic and highly detailed, yet it receives a lower CLIP score of 0.3 as compared to a score of 0.32 achieved by GaussianDreamer.
Similarly, for the \textit{Green Orc}, GSGen, despite having the Janus problem, scores the same as our method with a score of 0.31.
Therefore, we find the CLIP score to be unreliable as an evaluation criterion and thus conduct a user study to further establish the effectiveness of our method.
\end{comment}
\section{User Study}

\begin{figure}[!htp]
    \centering
    \includegraphics[width=0.4\textwidth]{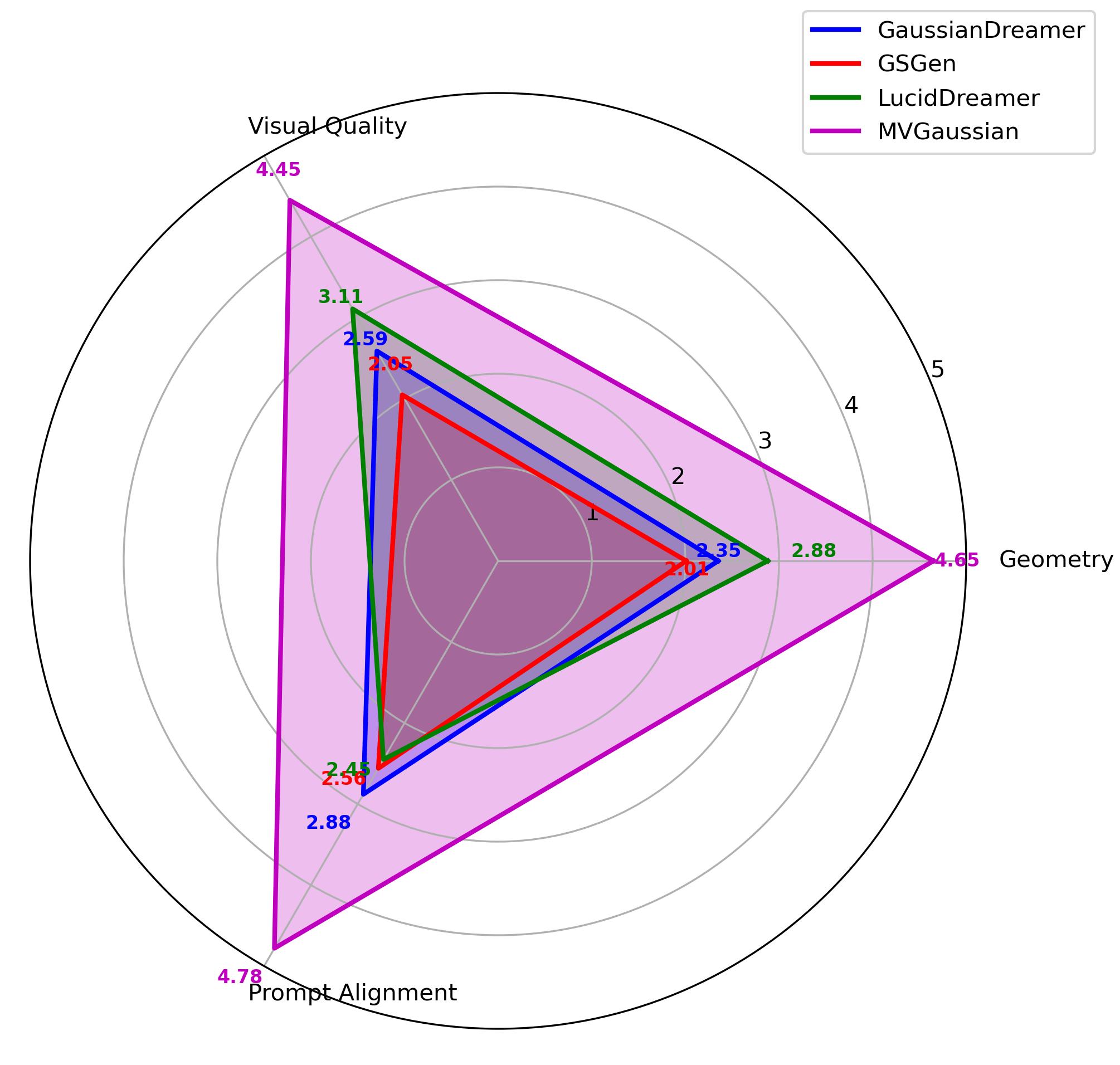}
    \caption{Evaluation of various aspects of the generated 3D content across different text-to-3D models based on human assessments.}
    \label{fig:human-eval}
\end{figure}

To further evaluate the performance of our method, we conducted a user survey with 42 participants as shown in Figure 
 \ref{fig:human-eval}. Each participant was asked to rate eight outputs generated by different text-to-3D models on a scale from 1 to 5 (higher is better). As shown in Table \ref{tab:comparison_human_eval}, the average human evaluation scores demonstrate that our method significantly outperforms other methods by a wide margin. Additionally, participants were asked to assess the models based on visual quality, geometry, and prompt alignment of the generated content. As illustrated in Figure \ref{fig:human-eval}, our method achieved superior scores across all metrics, with average ratings of $4.45$ for visual quality, $4.65$ for geometry, and $4.78$ for prompt alignment. In contrast, the next best-performing method, LucidDreamer, received average scores of $3.11$, $2.88$, and $2.45$, respectively. These human evaluation results underscore our method’s ability to produce more accurate and aesthetically pleasing 3D models, highlighting its effectiveness in overcoming the limitations of existing approaches.
Further, we demonstrate the effectiveness of our proposed method via ablations in the Appendix in Figure \ref{fig:2dvs3d}.

\section{Conclusion}

We present an intuitive and elegant method for high-quality text-to-3D renderings using depth maps without external supervision. Our approach employs the back-projection of screen space points to 3D for filtering gaussians and leverages multi-view diffusion guidance along with surface alignment to achieve superior results. This technique not only produces higher-quality renderings in significantly less time but also demonstrates robustness across diverse text prompts. Our method generates highly detailed renderings using Gaussian splatting in under half an hour, striking an optimal balance between quality and speed, unlike other methods. This establishes a rapid, SDS-based high-quality rendering scheme.

% \section{Acknowledgments}
% AAAI is especially grateful to Peter Patel Schneider for his work in implementing the original aaai.sty file, liberally using the ideas of other style hackers, including Barbara Beeton. We also acknowledge with thanks the work of George Ferguson for his guide to using the style and BibTeX files --- which has been incorporated into this document --- and Hans Guesgen, who provided several timely modifications, as well as the many others who have, from time to time, sent in suggestions on improvements to the AAAI style. We are especially grateful to Francisco Cruz, Marc Pujol-Gonzalez, and Mico Loretan for the improvements to the Bib\TeX{} and \LaTeX{} files made in 2020.

% The preparation of the \LaTeX{} and Bib\TeX{} files that implement these instructions was supported by Schlumberger Palo Alto Research, AT\&T Bell Laboratories, Morgan Kaufmann Publishers, The Live Oak Press, LLC, and AAAI Press. Bibliography style changes were added by Sunil Issar. \verb+\+pubnote was added by J. Scott Penberthy. George Ferguson added support for printing the AAAI copyright slug. Additional changes to aaai25.sty and aaai25.bst have been made by Francisco Cruz and Marc Pujol-Gonzalez.

% \bigskip
% \noindent Thank you for reading these instructions carefully. We look forward to receiving your electronic files!
\small
\bibliography{aaai25}

\begin{thebibliography}{32}
\providecommand{\natexlab}[1]{#1}

\bibitem[{Cao and Johnson(2023)}]{cao2023hexplane}
Cao, A.; and Johnson, J. 2023.
\newblock Hexplane: A fast representation for dynamic scenes.
\newblock In \emph{Proceedings of the IEEE/CVF Conference on Computer Vision and Pattern Recognition}, 130--141.

\bibitem[{Chen et~al.(2023{\natexlab{a}})Chen, Chen, Jiao, and Jia}]{chen2023fantasia3d}
Chen, R.; Chen, Y.; Jiao, N.; and Jia, K. 2023{\natexlab{a}}.
\newblock Fantasia3d: Disentangling geometry and appearance for high-quality text-to-3d content creation.
\newblock In \emph{Proceedings of the IEEE/CVF International Conference on Computer Vision}, 22246--22256.

\bibitem[{Chen et~al.(2023{\natexlab{b}})Chen, Chen, Jiao, and Jia}]{fantasia3d}
Chen, R.; Chen, Y.; Jiao, N.; and Jia, K. 2023{\natexlab{b}}.
\newblock Fantasia3D: Disentangling Geometry and Appearance for High-quality Text-to-3D Content Creation.
\newblock In \emph{Proceedings of the IEEE/CVF International Conference on Computer Vision (ICCV)}.

\bibitem[{Chen et~al.(2024)Chen, Wang, Wang, and Liu}]{gsgen}
Chen, Z.; Wang, F.; Wang, Y.; and Liu, H. 2024.
\newblock Text-to-3D using Gaussian Splatting.
\newblock arXiv:2309.16585.

\bibitem[{Dhariwal and Nichol(2021)}]{DiffvsGAN}
Dhariwal, P.; and Nichol, A. 2021.
\newblock Diffusion Models Beat GANs on Image Synthesis.
\newblock \emph{CoRR}, abs/2105.05233.

\bibitem[{Goodfellow et~al.(2014)Goodfellow, Pouget-Abadie, Mirza, Xu, Warde-Farley, Ozair, Courville, and Bengio}]{GAN}
Goodfellow, I.; Pouget-Abadie, J.; Mirza, M.; Xu, B.; Warde-Farley, D.; Ozair, S.; Courville, A.; and Bengio, Y. 2014.
\newblock Generative Adversarial Nets.
\newblock In Ghahramani, Z.; Welling, M.; Cortes, C.; Lawrence, N.; and Weinberger, K., eds., \emph{Advances in Neural Information Processing Systems}, volume~27. Curran Associates, Inc.

\bibitem[{Gu{\'e}don and Lepetit(2023)}]{sugar}
Gu{\'e}don, A.; and Lepetit, V. 2023.
\newblock Sugar: Surface-aligned gaussian splatting for efficient 3d mesh reconstruction and high-quality mesh rendering.
\newblock \emph{arXiv preprint arXiv:2311.12775}.

\bibitem[{Ho, Jain, and Abbeel(2020{\natexlab{a}})}]{ho2020denoising}
Ho, J.; Jain, A.; and Abbeel, P. 2020{\natexlab{a}}.
\newblock Denoising diffusion probabilistic models.
\newblock \emph{Advances in neural information processing systems}, 33: 6840--6851.

\bibitem[{Ho, Jain, and Abbeel(2020{\natexlab{b}})}]{DDPM}
Ho, J.; Jain, A.; and Abbeel, P. 2020{\natexlab{b}}.
\newblock Denoising Diffusion Probabilistic Models.
\newblock \emph{CoRR}, abs/2006.11239.

\bibitem[{Ho and Salimans(2021)}]{CFG}
Ho, J.; and Salimans, T. 2021.
\newblock Classifier-Free Diffusion Guidance.
\newblock In \emph{NeurIPS 2021 Workshop on Deep Generative Models and Downstream Applications}.

\bibitem[{Huang et~al.(2024)Huang, Yu, Chen, Geiger, and Gao}]{huang20242d}
Huang, B.; Yu, Z.; Chen, A.; Geiger, A.; and Gao, S. 2024.
\newblock 2d gaussian splatting for geometrically accurate radiance fields.
\newblock In \emph{ACM SIGGRAPH 2024 Conference Papers}, 1--11.

\bibitem[{Kendall, Gal, and Cipolla(2017)}]{UncertaintyLoss}
Kendall, A.; Gal, Y.; and Cipolla, R. 2017.
\newblock Multi-Task Learning Using Uncertainty to Weigh Losses for Scene Geometry and Semantics.
\newblock \emph{CoRR}, abs/1705.07115.

\bibitem[{Kerbl et~al.(2023)Kerbl, Kopanas, Leimk{\"u}hler, and Drettakis}]{3DGS}
Kerbl, B.; Kopanas, G.; Leimk{\"u}hler, T.; and Drettakis, G. 2023.
\newblock 3D Gaussian Splatting for Real-Time Radiance Field Rendering.
\newblock \emph{ACM Transactions on Graphics}, 42(4).

\bibitem[{Liang et~al.(2023)Liang, Yang, Lin, Li, Xu, and Chen}]{luciddreamer}
Liang, Y.; Yang, X.; Lin, J.; Li, H.; Xu, X.; and Chen, Y. 2023.
\newblock LucidDreamer: Towards High-Fidelity Text-to-3D Generation via Interval Score Matching.
\newblock arXiv:2311.11284.

\bibitem[{Lin et~al.(2023)Lin, Gao, Tang, Takikawa, Zeng, Huang, Kreis, Fidler, Liu, and Lin}]{lin2023magic3d}
Lin, C.-H.; Gao, J.; Tang, L.; Takikawa, T.; Zeng, X.; Huang, X.; Kreis, K.; Fidler, S.; Liu, M.-Y.; and Lin, T.-Y. 2023.
\newblock Magic3d: High-resolution text-to-3d content creation.
\newblock In \emph{Proceedings of the IEEE/CVF Conference on Computer Vision and Pattern Recognition}, 300--309.

\bibitem[{Liu et~al.(2023)Liu, Wu, Van~Hoorick, Tokmakov, Zakharov, and Vondrick}]{liu2023zero}
Liu, R.; Wu, R.; Van~Hoorick, B.; Tokmakov, P.; Zakharov, S.; and Vondrick, C. 2023.
\newblock Zero-1-to-3: Zero-shot one image to 3d object.
\newblock In \emph{Proceedings of the IEEE/CVF International Conference on Computer Vision}, 9298--9309.

\bibitem[{Mildenhall et~al.(2020)Mildenhall, Srinivasan, Tancik, Barron, Ramamoorthi, and Ng}]{NERF}
Mildenhall, B.; Srinivasan, P.~P.; Tancik, M.; Barron, J.~T.; Ramamoorthi, R.; and Ng, R. 2020.
\newblock NeRF: Representing Scenes as Neural Radiance Fields for View Synthesis.
\newblock \emph{CoRR}, abs/2003.08934.

\bibitem[{Nichol et~al.(2022)Nichol, Jun, Dhariwal, Mishkin, and Chen}]{nichol2022point}
Nichol, A.; Jun, H.; Dhariwal, P.; Mishkin, P.; and Chen, M. 2022.
\newblock Point-e: A system for generating 3d point clouds from complex prompts.
\newblock \emph{arXiv preprint arXiv:2212.08751}.

\bibitem[{Poole et~al.(2022)Poole, Jain, Barron, and Mildenhall}]{dreamfusion}
Poole, B.; Jain, A.; Barron, J.~T.; and Mildenhall, B. 2022.
\newblock DreamFusion: Text-to-3D using 2D Diffusion.
\newblock \emph{arXiv}.

\bibitem[{Radford et~al.(2021)Radford, Kim, Hallacy, Ramesh, Goh, Agarwal, Sastry, Askell, Mishkin, Clark et~al.}]{radford2021learning}
Radford, A.; Kim, J.~W.; Hallacy, C.; Ramesh, A.; Goh, G.; Agarwal, S.; Sastry, G.; Askell, A.; Mishkin, P.; Clark, J.; et~al. 2021.
\newblock Learning transferable visual models from natural language supervision.
\newblock In \emph{International conference on machine learning}, 8748--8763. PMLR.

\bibitem[{Ramesh et~al.(2022)Ramesh, Dhariwal, Nichol, Chu, and Chen}]{ramesh2022hierarchical}
Ramesh, A.; Dhariwal, P.; Nichol, A.; Chu, C.; and Chen, M. 2022.
\newblock Hierarchical text-conditional image generation with clip latents.
\newblock \emph{arXiv preprint arXiv:2204.06125}, 1(2): 3.

\bibitem[{Ramesh et~al.(2021)Ramesh, Pavlov, Goh, Gray, Voss, Radford, Chen, and Sutskever}]{ramesh2021zero}
Ramesh, A.; Pavlov, M.; Goh, G.; Gray, S.; Voss, C.; Radford, A.; Chen, M.; and Sutskever, I. 2021.
\newblock Zero-shot text-to-image generation.
\newblock In \emph{International conference on machine learning}, 8821--8831. Pmlr.

\bibitem[{Rombach et~al.(2022)Rombach, Blattmann, Lorenz, Esser, and Ommer}]{rombach2022high}
Rombach, R.; Blattmann, A.; Lorenz, D.; Esser, P.; and Ommer, B. 2022.
\newblock High-resolution image synthesis with latent diffusion models.
\newblock In \emph{Proceedings of the IEEE/CVF conference on computer vision and pattern recognition}, 10684--10695.

\bibitem[{Saharia et~al.(2022)Saharia, Chan, Saxena, Li, Whang, Denton, Ghasemipour, Gontijo~Lopes, Karagol~Ayan, Salimans et~al.}]{imagen}
Saharia, C.; Chan, W.; Saxena, S.; Li, L.; Whang, J.; Denton, E.~L.; Ghasemipour, K.; Gontijo~Lopes, R.; Karagol~Ayan, B.; Salimans, T.; et~al. 2022.
\newblock Photorealistic text-to-image diffusion models with deep language understanding.
\newblock \emph{Advances in neural information processing systems}, 35: 36479--36494.

\bibitem[{Shen et~al.(2021)Shen, Gao, Yin, Liu, and Fidler}]{shen2021deep}
Shen, T.; Gao, J.; Yin, K.; Liu, M.-Y.; and Fidler, S. 2021.
\newblock Deep marching tetrahedra: a hybrid representation for high-resolution 3d shape synthesis.
\newblock \emph{Advances in Neural Information Processing Systems}, 34: 6087--6101.

\bibitem[{Shi et~al.(2024)Shi, Wang, Ye, Mai, Li, and Yang}]{mvdream}
Shi, Y.; Wang, P.; Ye, J.; Mai, L.; Li, K.; and Yang, X. 2024.
\newblock {MVD}ream: Multi-view Diffusion for 3D Generation.
\newblock In \emph{The Twelfth International Conference on Learning Representations}.

\bibitem[{Song, Meng, and Ermon(2020)}]{song2020denoising}
Song, J.; Meng, C.; and Ermon, S. 2020.
\newblock Denoising diffusion implicit models.
\newblock \emph{arXiv preprint arXiv:2010.02502}.

\bibitem[{Tang et~al.(2024)Tang, Ren, Zhou, Liu, and Zeng}]{dreamgaussian}
Tang, J.; Ren, J.; Zhou, H.; Liu, Z.; and Zeng, G. 2024.
\newblock DreamGaussian: Generative Gaussian Splatting for Efficient 3D Content Creation.
\newblock In \emph{The Twelfth International Conference on Learning Representations}.

\bibitem[{Wang et~al.(2023{\natexlab{a}})Wang, Du, Li, Yeh, and Shakhnarovich}]{wang2023score}
Wang, H.; Du, X.; Li, J.; Yeh, R.~A.; and Shakhnarovich, G. 2023{\natexlab{a}}.
\newblock Score jacobian chaining: Lifting pretrained 2d diffusion models for 3d generation.
\newblock In \emph{Proceedings of the IEEE/CVF Conference on Computer Vision and Pattern Recognition}, 12619--12629.

\bibitem[{Wang et~al.(2023{\natexlab{b}})Wang, Lu, Wang, Bao, Li, Su, and Zhu}]{prolificdreamer}
Wang, Z.; Lu, C.; Wang, Y.; Bao, F.; Li, C.; Su, H.; and Zhu, J. 2023{\natexlab{b}}.
\newblock ProlificDreamer: High-Fidelity and Diverse Text-to-3D Generation with Variational Score Distillation.
\newblock In \emph{Thirty-seventh Conference on Neural Information Processing Systems}.

\bibitem[{Yi et~al.(2024)Yi, Fang, Wang, Wu, Xie, Zhang, Liu, Tian, and Wang}]{gaussiandreamer}
Yi, T.; Fang, J.; Wang, J.; Wu, G.; Xie, L.; Zhang, X.; Liu, W.; Tian, Q.; and Wang, X. 2024.
\newblock GaussianDreamer: Fast Generation from Text to 3D Gaussians by Bridging 2D and 3D Diffusion Models.
\newblock In \emph{CVPR}.

\bibitem[{Zhang, Rao, and Agrawala(2023)}]{zhang2023adding}
Zhang, L.; Rao, A.; and Agrawala, M. 2023.
\newblock Adding conditional control to text-to-image diffusion models.
\newblock In \emph{Proceedings of the IEEE/CVF International Conference on Computer Vision}, 3836--3847.

\end{thebibliography}

\clearpage

\section{Appendix}
\appendix

\subsection{Quantitative Evaluation}

\begin{table*}[!htp]
\centering
\scalebox{0.7}{
\begin{tabular}{|p{15cm}|c|c|c|c|}
\hline
\textbf{Prompt} & \multicolumn{4}{c|}{\textbf{CLIP score}} \\ \cline{2-5} 
 & \textbf{GaussianDreamer} & \textbf{GSGEN} & \textbf{LucidDreamer} & \textbf{Ours} \\ \hline
``A blue jay sitting on a willow basket of macarons" & 0.28 & 0.27 & 0.33 & \textbf{0.34} \\ \hline
``A flying dragon, highly detailed, realistic, majestic.`` & 0.3 & 0.3 & 0.3 & \textbf{0.32} \\ \hline
``An armored green-skin orc warrior riding a vicious hog." & 0.29 & \textbf{0.31} & 0.29 & \textbf{0.31} \\ \hline
``A forbidden castle high up in the mountains." & \textbf{0.32} & 0.25 & 0.27 & 0.30 \\ \hline
``A peacock standing on a surfing board, highly detailed, majestic." & 0.31 & 0.29 & 0.29 & \textbf{0.33} \\ \hline
``Jack Sparrow wearing sunglasses, head, photorealistic, 8k, HD, raw." & 0.27 & \textbf{0.33} & 0.29 & 0.31 \\ \hline
``Medieval soldier with shield and sword, fantasy, game, character, highly detailed, photorealistic, 4K, HD" & 0.25 & 0.29 & 0.26 & \textbf{0.31} \\ \hline
``A 3D model of an adorable cottage with a thatched roof" & 0.32 & 0.31 & \textbf{0.34} & 0.31 \\ \hline
\end{tabular}
}
\caption{Comparison of different methods based on the given prompt. The CLIP score is computed for 15 views generated from the 3D model of each method, then averaged. We also compute the number of Gaussians for the naive method, which does not use the surface densification proposed by our method.}

\label{tab:comparison}
\end{table*}

% \begin{table}[!htp]
% \centering
% \scalebox{0.65}{
% \begin{tabular}{|p{4cm}|c|c|c|c|}
% \hline
% \textbf{Prompt} & \multicolumn{4}{c|}{\textbf{CLIP score}}\\ \cline{2-5} 
%  & \textbf{GaussianDreamer} & \textbf{Gsgen} & \textbf{LucidDreamer} & \textbf{Ours}  \\ \hline
% "A \textbf{blue jay}..." & 0.28 & 0.27 & 0.33 & \textbf{0.34}  \\ \hline
% "A flying \textbf{dragon}..." & 0.3 & 0.3 & 0.3 & \textbf{0.32}  \\ \hline
% "An \textbf{orc warrior}..." & 0.29 & \textbf{0.31} & 0.29 & \textbf{0.31} \\ \hline
% "A \textbf{castle}..." & \textbf{0.32} & 0.25 & 0.27 & 0.3  \\ \hline
% "A \textbf{peacock}..." & 0.31 & 0.29 & 0.29 & \textbf{0.33}  \\ \hline
% "\textbf{Jack Sparrow}..." & 0.27 & \textbf{0.33} & 0.29 & 0.31 \\ \hline
% "\textbf{Medieval soldier}..." & 0.25 & 0.29 & 0.26 & \textbf{0.31}  \\ \hline
% "A  \textbf{cottage}..." & 0.32 & 0.31 & \textbf{0.34} & 0.31  \\ \hline
% \end{tabular}}
% \caption{Comparison of different methods based on the given prompt. The CLIP score is computed for 15 views generated from the 3D model of each method, then averaged. We also compute the number of Gaussians for the naive method, which does not use the surface densification proposed by our method.}

% \label{tab:comparison}
% \end{table}
% \vspace{-5pt}

% \section{Ablation}

% \section{Training Details}
%%% detailed training details and params for reproducibility checklist
In Table \ref{tab:comparison}, we compare the performance of our method against GaussianDreamer \cite{gaussiandreamer}, GSGEN \cite{gsgen}, and LucidDreamer \cite{luciddreamer} using the CLIP score, averaged over 15 views generated from 3D models for each prompt shown in \textbf{Figure 2 of the main manuscript}. Our method consistently achieves the highest or near-highest scores across various prompts, such as \textbf{\textit{Blue jay}}, \textbf{\textit{Peacock}}, and \textbf{\textit{Dragon}}, indicating superior text-image alignment. However, despite the superior visual quality, the CLIP scores do not show a significant improvement for all prompts, and in some cases (\textbf{\textit{Castle}}, \textbf{\textit{Jack Sparrow}}, \textbf{\textit{Cottage}}), our CLIP scores are even lower. This discrepancy arises because the CLIP score, while useful for measuring 2D image-text alignment, is not a reliable metric for evaluating the performance of text-to-3D models. The CLIP score does not fully capture the fidelity, coherence, or geometric accuracy of the 3D models across different views, leading to potential underestimation of the quality improvements introduced by our method; for, e.g., the Jacksparrow model, despite having Janus problem in GSGen scores higher than ours. Therefore, additional metrics beyond the CLIP score, such as human evaluation in the form of user studies as we conducted, are necessary to thoroughly assess the overall quality of text-to-3D model generation.

\subsection{Ablation studies}

\begin{figure}[!htp]
        \centering
        \setlength{\tabcolsep}{-5pt}
        \scalebox{0.55}{
        \begin{tabular}{cccc}
        Without flattening & With flattening & Without flattening & With flattening  \\
         \& opacity regularization &  \& opacity regularization  & \& opacity regularization &  \& opacity regularization \\
        \multicolumn{2}{c}{"A forbidden castle high up in the mountains"} & 
            \multicolumn{2}{c}{"A 3D model of an adorable cottage with a thatched roof"} \\
           \includegraphics[width=0.5\linewidth]{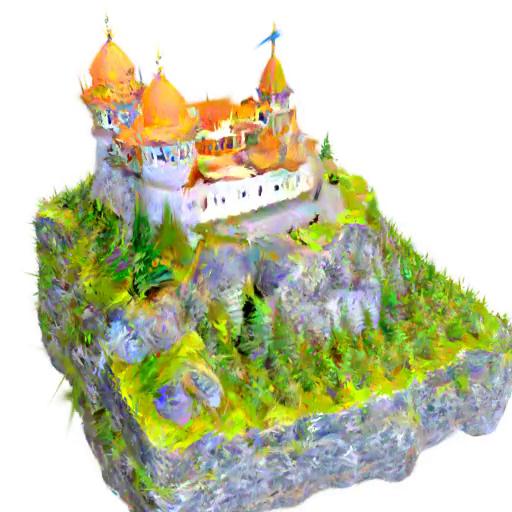}  & 
            \includegraphics[width=0.5\linewidth]{gscraft_no_regularization/castle/ours_losses/view_6.jpg} &

            \includegraphics[width=0.45\linewidth]{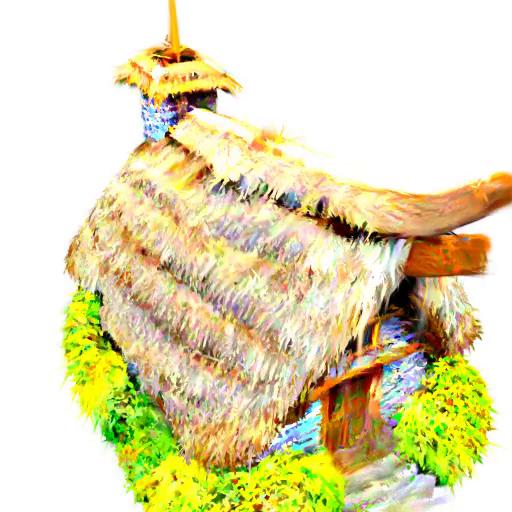} & 
            \includegraphics[width=0.55\linewidth]{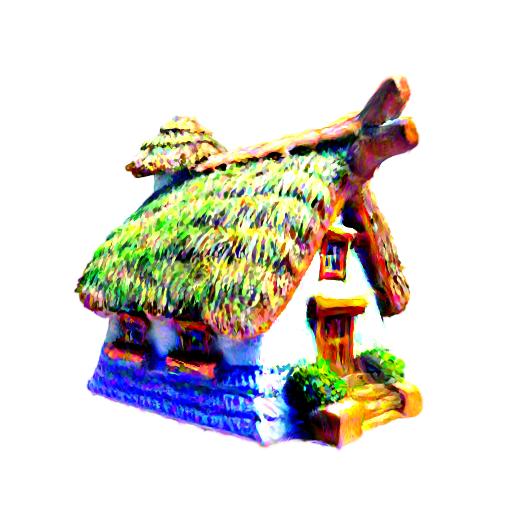} 
            \\
            \multicolumn{2}{c}{"Jack Sparrow wearing sunglasses, } & 
            \multicolumn{2}{c}{“A DSLR photo of the Imperial State Crown of England."}\\
             \multicolumn{2}{c}{head, photorealistic, 8K, HD, raw"} & \\
             \includegraphics[width=0.45\linewidth]
             {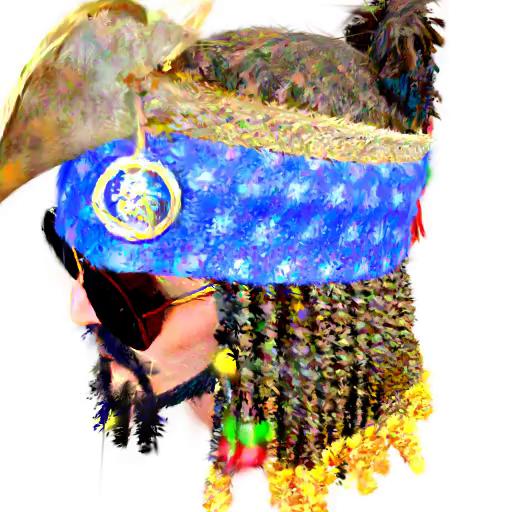} & 
            \includegraphics[width=0.4\linewidth]{gscraft_no_regularization/jack_sparrow/our_losses/jack_phu_view_7.jpg} 
            &
              \includegraphics[width=0.45\linewidth]{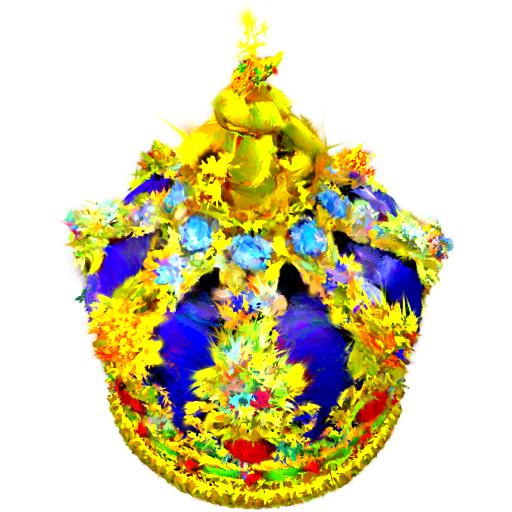} & 
            \includegraphics[width=0.55\linewidth]{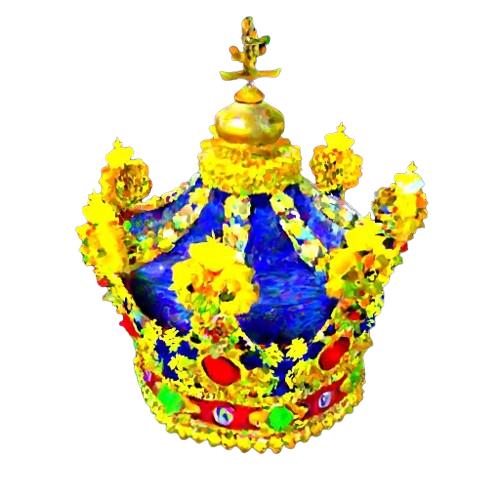}         
             
        \end{tabular}}

    \caption{Images with and without additional losses}
    \label{fig:loss_ablation}
\end{figure}

The Figure \ref{fig:loss_ablation} demonstrates the impact of applying regularization terms. The images on the left in each pair are generated without these regularization losses, while the images on the right incorporate them. The comparison highlights that models utilizing these regularization techniques produce more refined and visually compelling results, characterized by clearer details and reduced artifacts. For instance, in the \textbf{\textit{Castle}} example, the model with regularization achieves a more cohesive structure and vibrant color palette. Likewise, the \textbf{\textit{Crown}} with regularization displays a more polished and realistic appearance, with additional details and enhanced structural elements. The \textbf{\textit{Cottage}} and \textbf{\textit{Jack Sparrow}} examples also benefit from regularization, showing sharper details and more accurate textures, leading to a more realistic and appealing visual representation.
 
Further, in Figure \ref{fig:2dvs3d} in the supplementary, we illustrate a comparison with the incorporation of 2DGS \cite{huang20242d} and demonstrate that incorporation of our proposed regularization terms leads to sharper renderings and improved geometric structures, exhibiting both smoother and more detailed features.

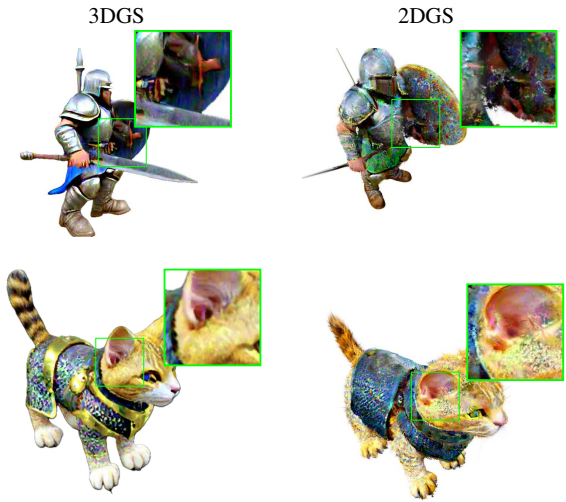
\begin{figure}[!htp]
    \centering
    \scalebox{0.85}{
    \begin{tabular}{cc}
         3DGS & 2DGS   \\

          \begin{tikzpicture}[spy using outlines={rectangle,magnification=2,size=1.5cm}]
				\node {\includegraphics[width=0.35\linewidth]{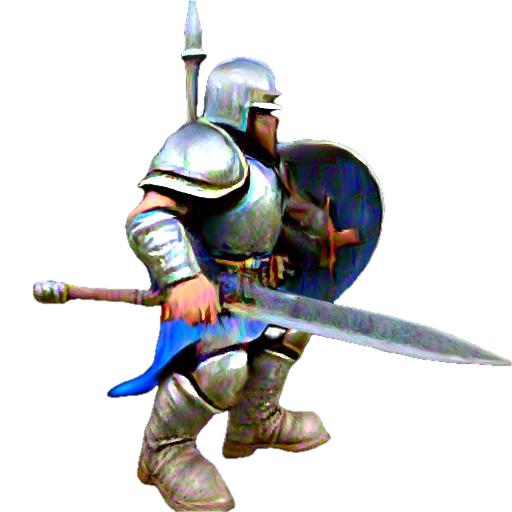}};
				%\node[] at (1, 0) {};
				%\spy[color=green] on (-0.6,-0.6) in node [left] at (-1, 1);
				\node[] at (1, 0) {};
				\spy[color=green] on (0.3,0.) in node [right] at (0.5, 1);
			\end{tikzpicture} &
        \begin{tikzpicture}[spy using outlines={rectangle,magnification=2,size=1.5cm}]
				\node {\includegraphics[width=0.35\linewidth]{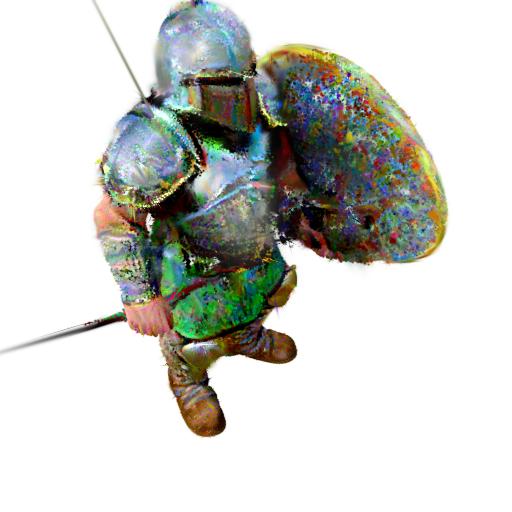}};
				%\node[] at (1, 0) {};
				%\spy[color=green] on (-0.6,-0.6) in node [left] at (-1, 1);
				\node[] at (1, 0) {};
				\spy[color=green] on (0.3,0.3) in node [right] at (1, 1);
			\end{tikzpicture}
         
          \\

        \begin{tikzpicture}[spy using outlines={rectangle,magnification=2,size=1.5cm}]
				\node { \includegraphics[width=0.45\linewidth]{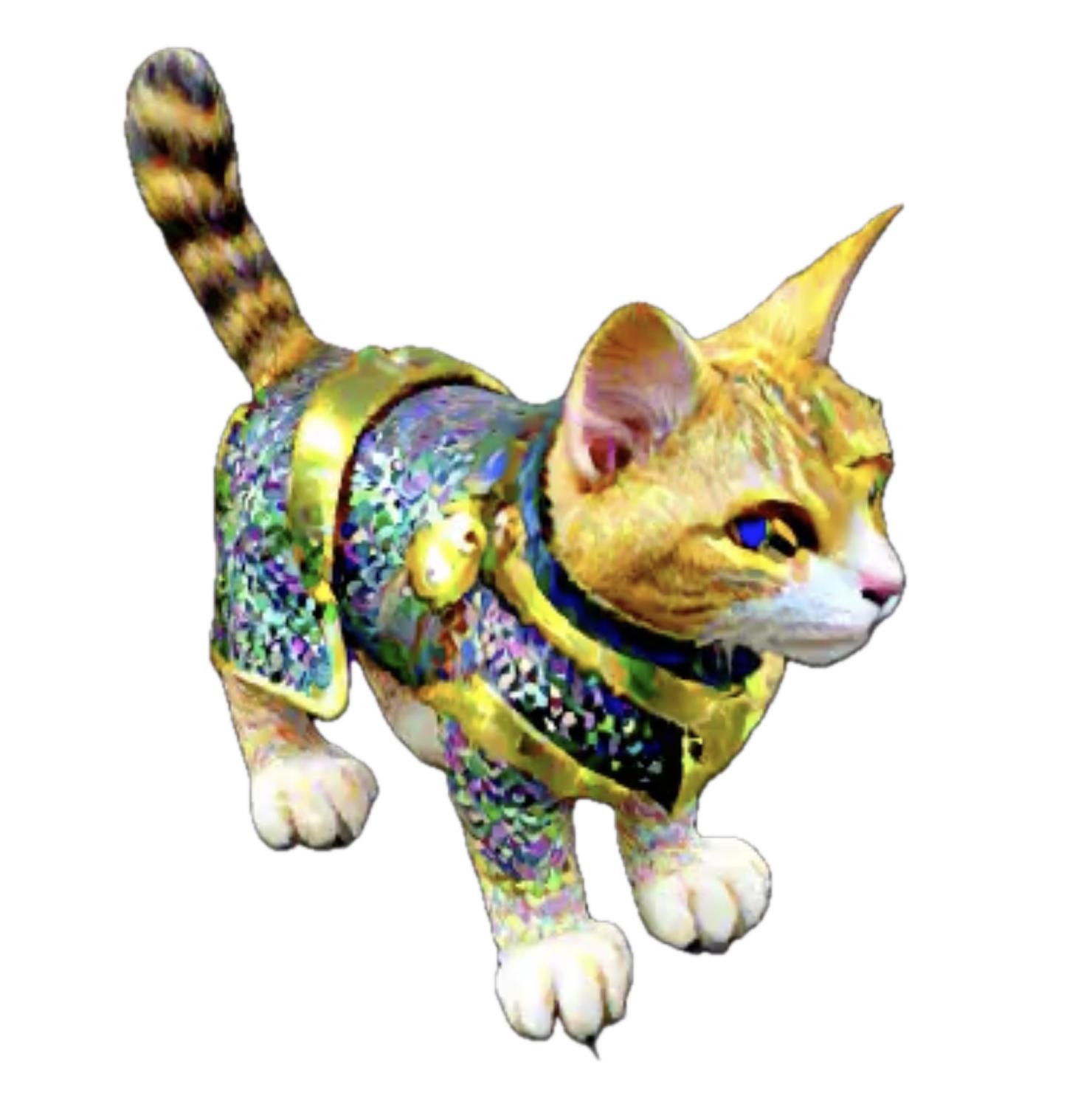}};
				%\node[] at (1, 0) {};
				%\spy[color=green] on (-0.6,-0.6) in node [left] at (-1, 1);
				\node[] at (1, 0) {};
				\spy[color=green] on (0.3,0.3) in node [right] at (1, 1);
			\end{tikzpicture}
         
         &
         \begin{tikzpicture}[spy using outlines={rectangle,magnification=2,size=1.5cm}]
				\node { \includegraphics[width=0.40\linewidth]{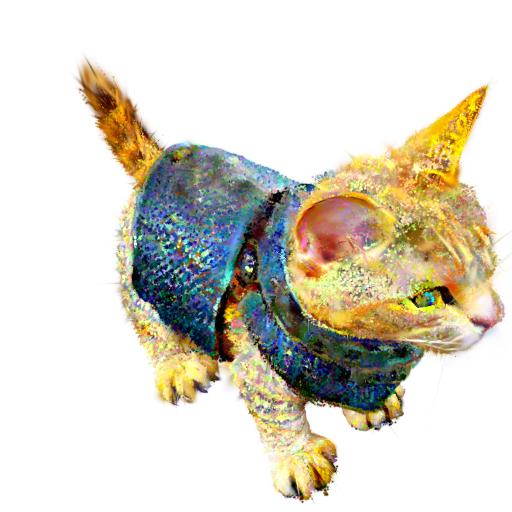}};
				%\node[] at (1, 0) {};
				%\spy[color=green] on (-0.6,-0.6) in node [left] at (-1, 1);
				\node[] at (1, 0) {};
				\spy[color=green] on (0.5,0.) in node [right] at (1, 1);
			\end{tikzpicture}
    \end{tabular}}
    \caption{Artifacts observed when densification is based on 2D Gaussian splatting (2DGS) compared to 3D Gaussian splatting (3DGS). The 2DGS method results in noticeable artifacts, particularly around regions like the hilt of the weapon and the tail of the cat, leading to a loss of details.}
    \label{fig:2dvs3d}
\end{figure}

\subsection{More Qualitative Comparisons}

\begin{figure*}[!htp]
    \centering
    \setlength{\tabcolsep}{-1pt}
    \scalebox{0.9}{
    \begin{tabular}{cccccccc}
        \multicolumn{2}{c}{\textbf{GaussianDreamer}} & \multicolumn{2}{c}{\textbf{LucidDreamer}} & 
        \multicolumn{2}{c}{\textbf{GSGen}} & \multicolumn{2}{c}{\textbf{MVGaussian}(ours)} \\
        \multicolumn{8}{c}{"Michelangelo style statue of a dog reading news on a cellphone."} \\
        \includegraphics[width=0.15\linewidth]{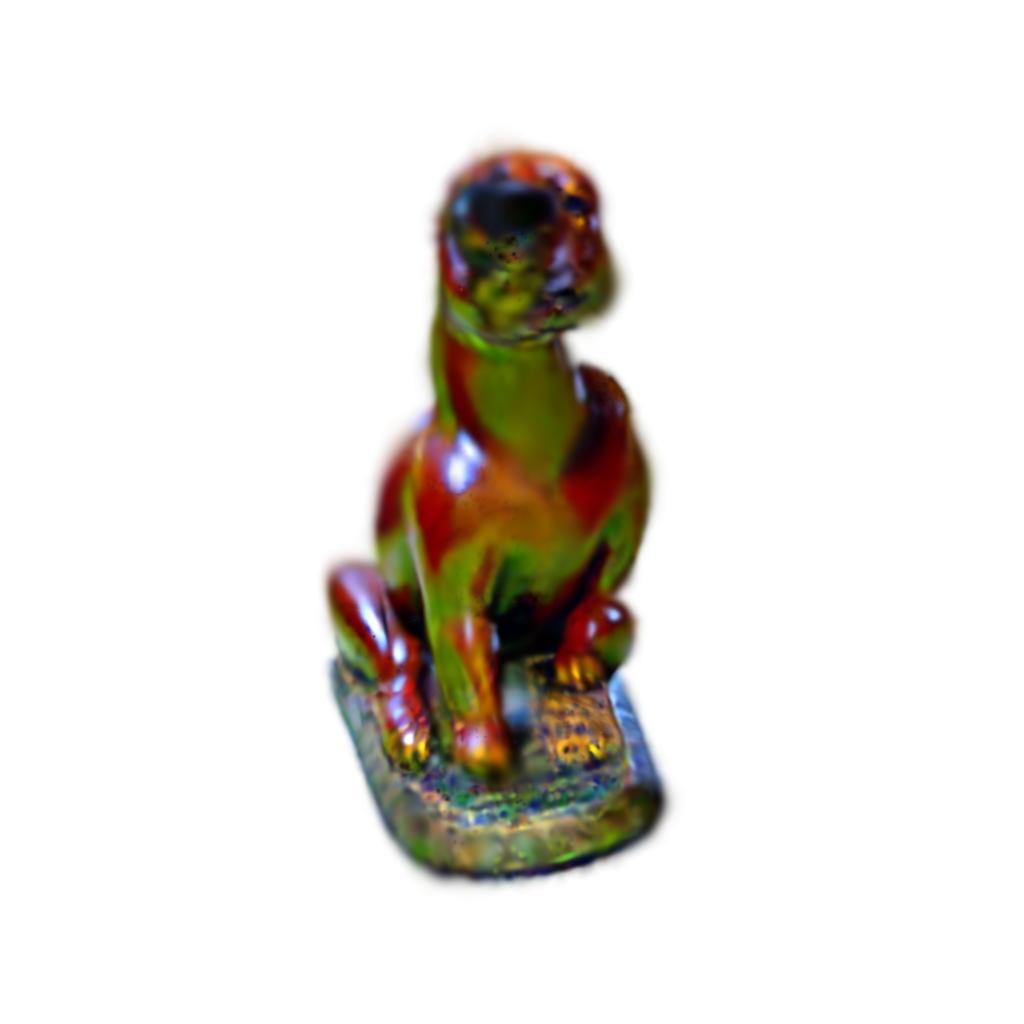} &
        \includegraphics[width=0.15\linewidth]{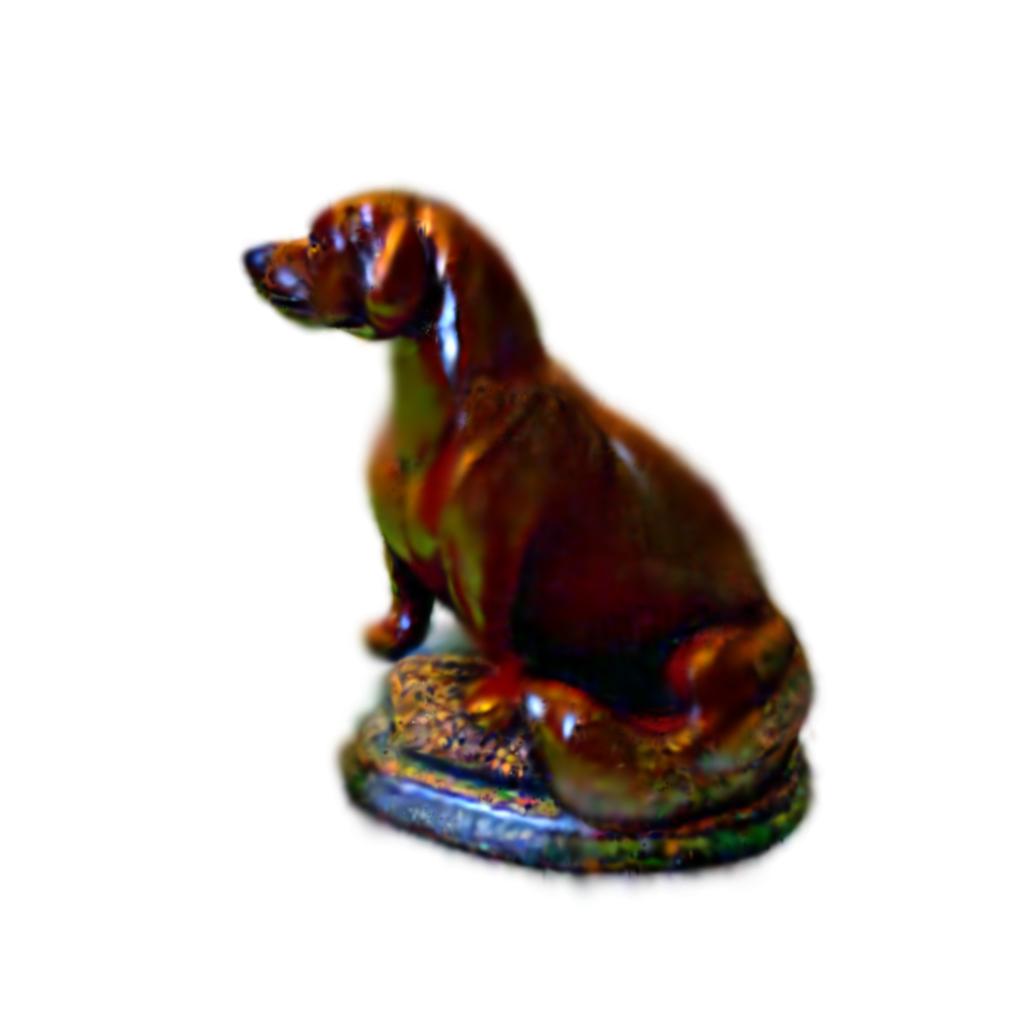} &
        \includegraphics[width=0.13\linewidth]{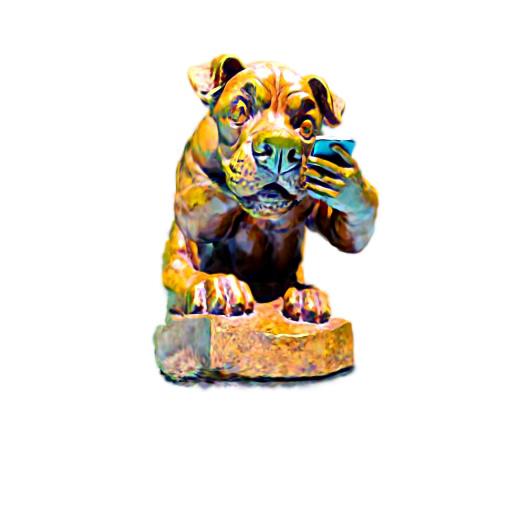} &
        \includegraphics[width=0.13\linewidth]{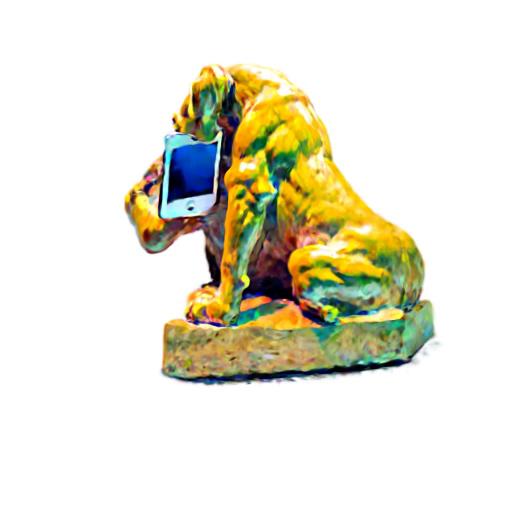} &
        \includegraphics[width=0.12\linewidth]{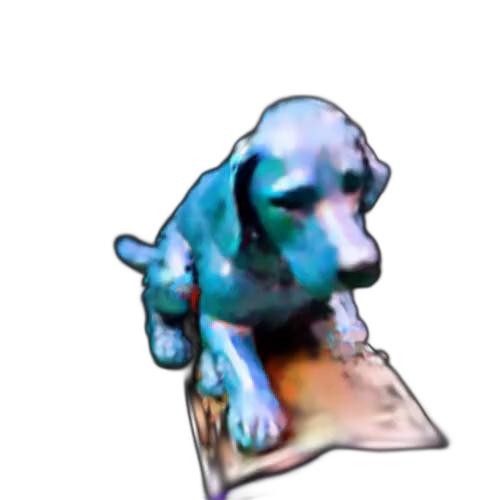} &
        \includegraphics[width=0.11\linewidth]{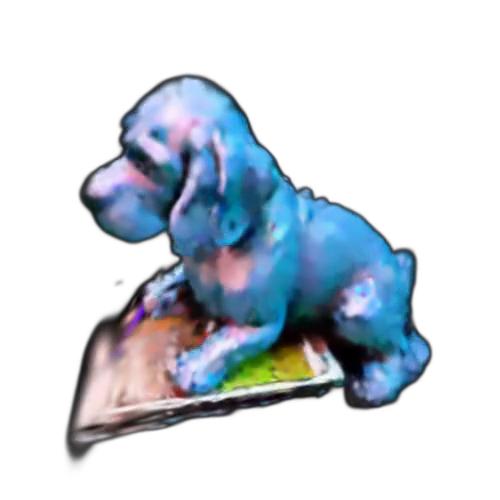} &  
        \includegraphics[width=0.12\linewidth]{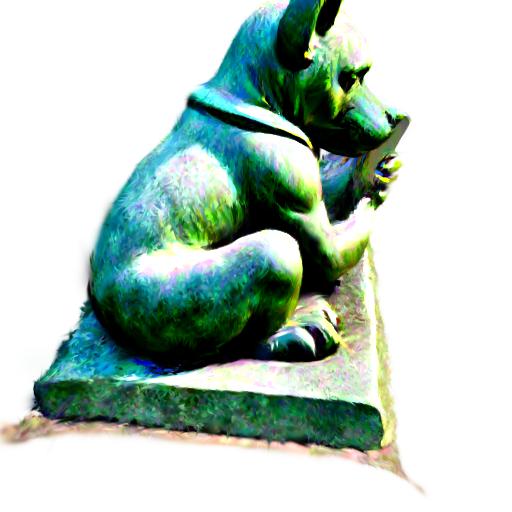} &
        \includegraphics[width=0.12\linewidth]{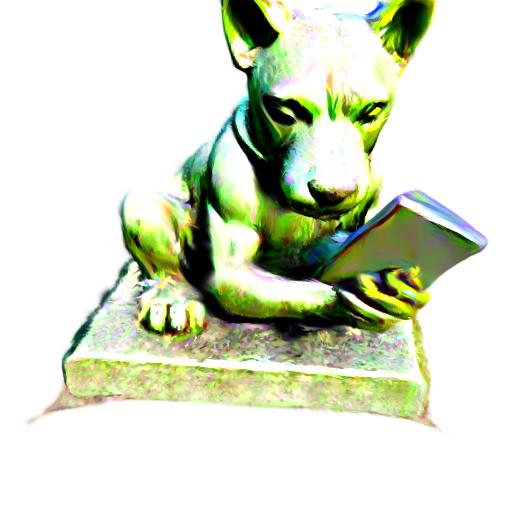} \\
        \multicolumn{8}{c}{"Airplane, fighter, steampunk style, ultra realistic, 4k, HD"} \\
        \includegraphics[width=0.14\linewidth]{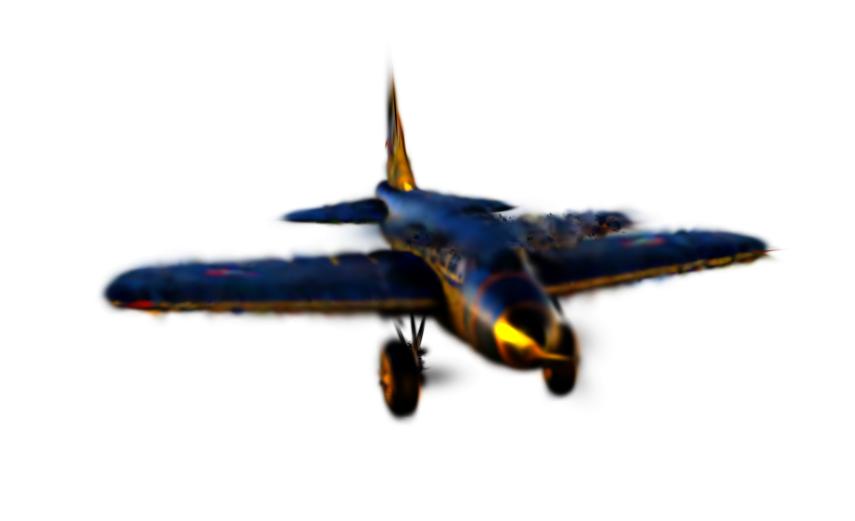} &  
        \includegraphics[width=0.14\linewidth]{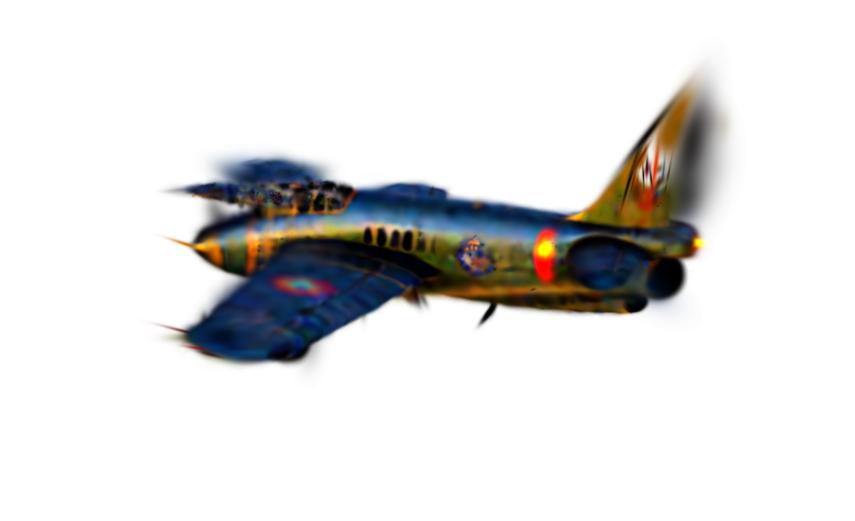} &
        \includegraphics[width=0.14\linewidth, trim={0cm 0cm 2cm 0cm},clip]{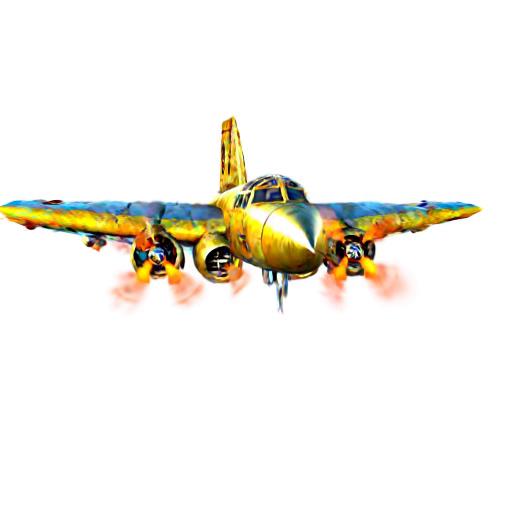} &
        \includegraphics[width=0.14\linewidth]{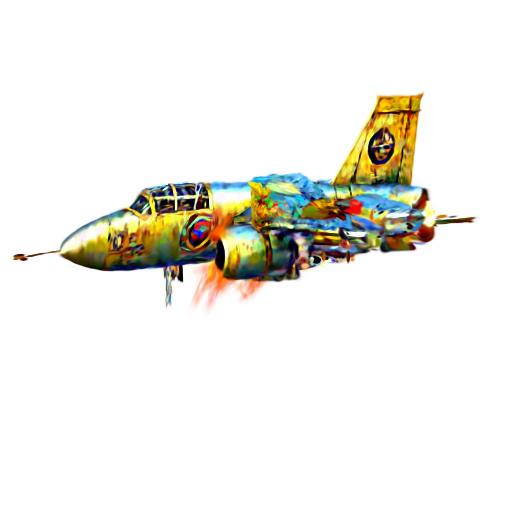} &  
        \includegraphics[width=0.09\linewidth]{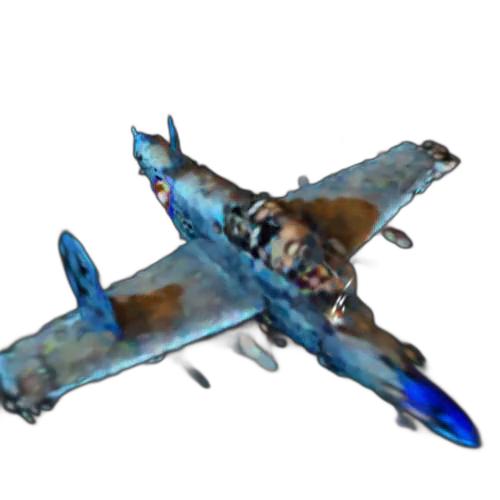} &  
        \includegraphics[width=0.09\linewidth]{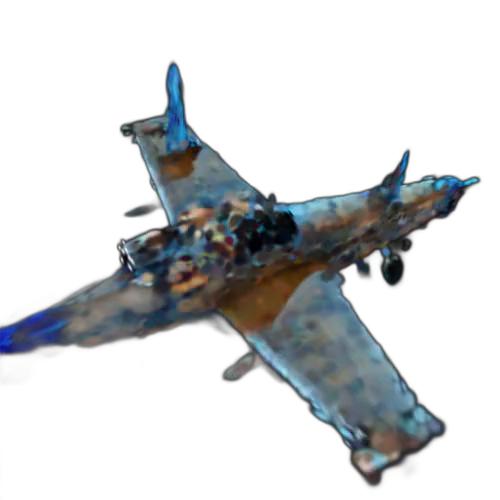}  &  
        \includegraphics[width=0.12\linewidth]{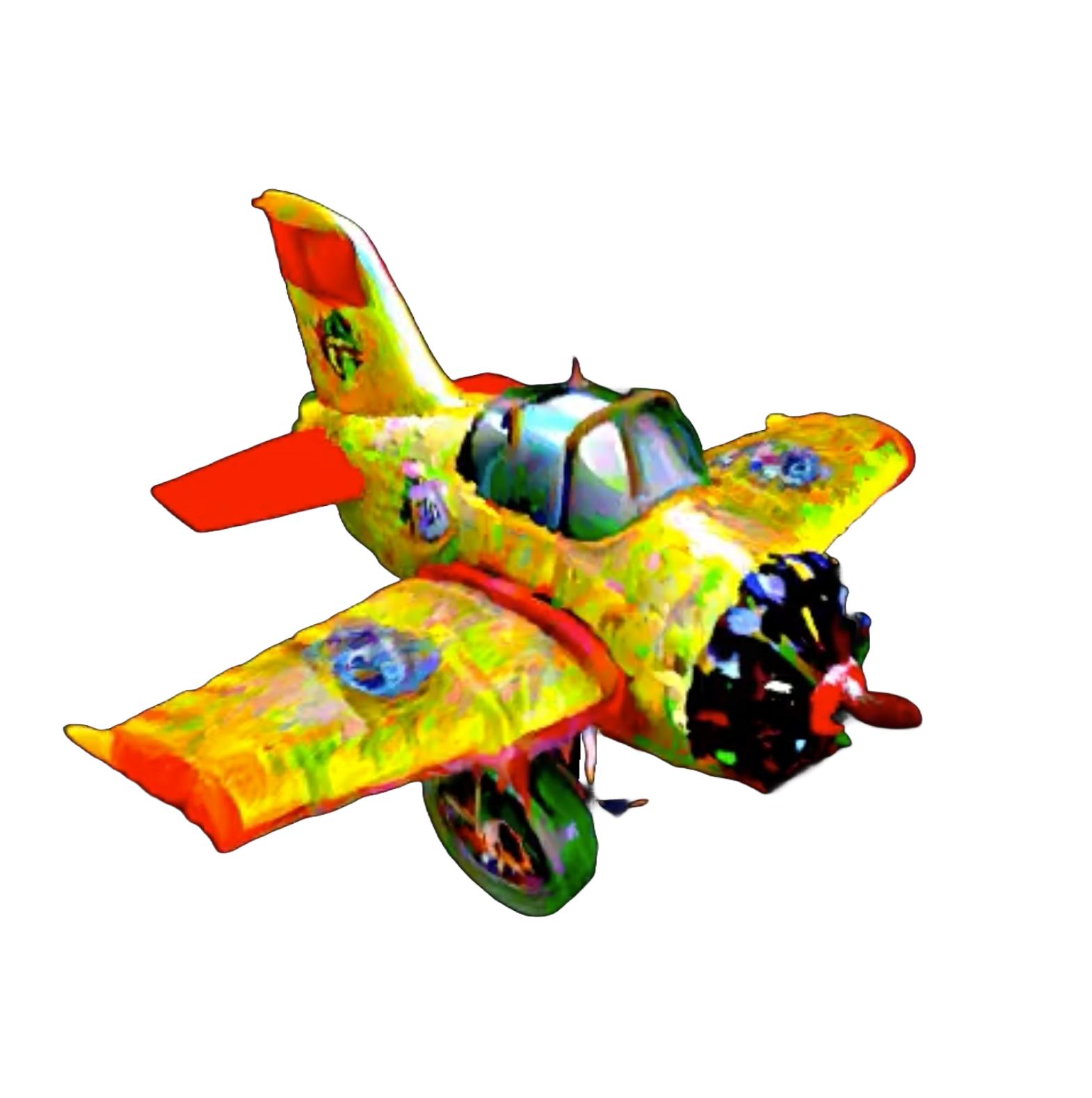} &
        \includegraphics[width=0.12\linewidth]{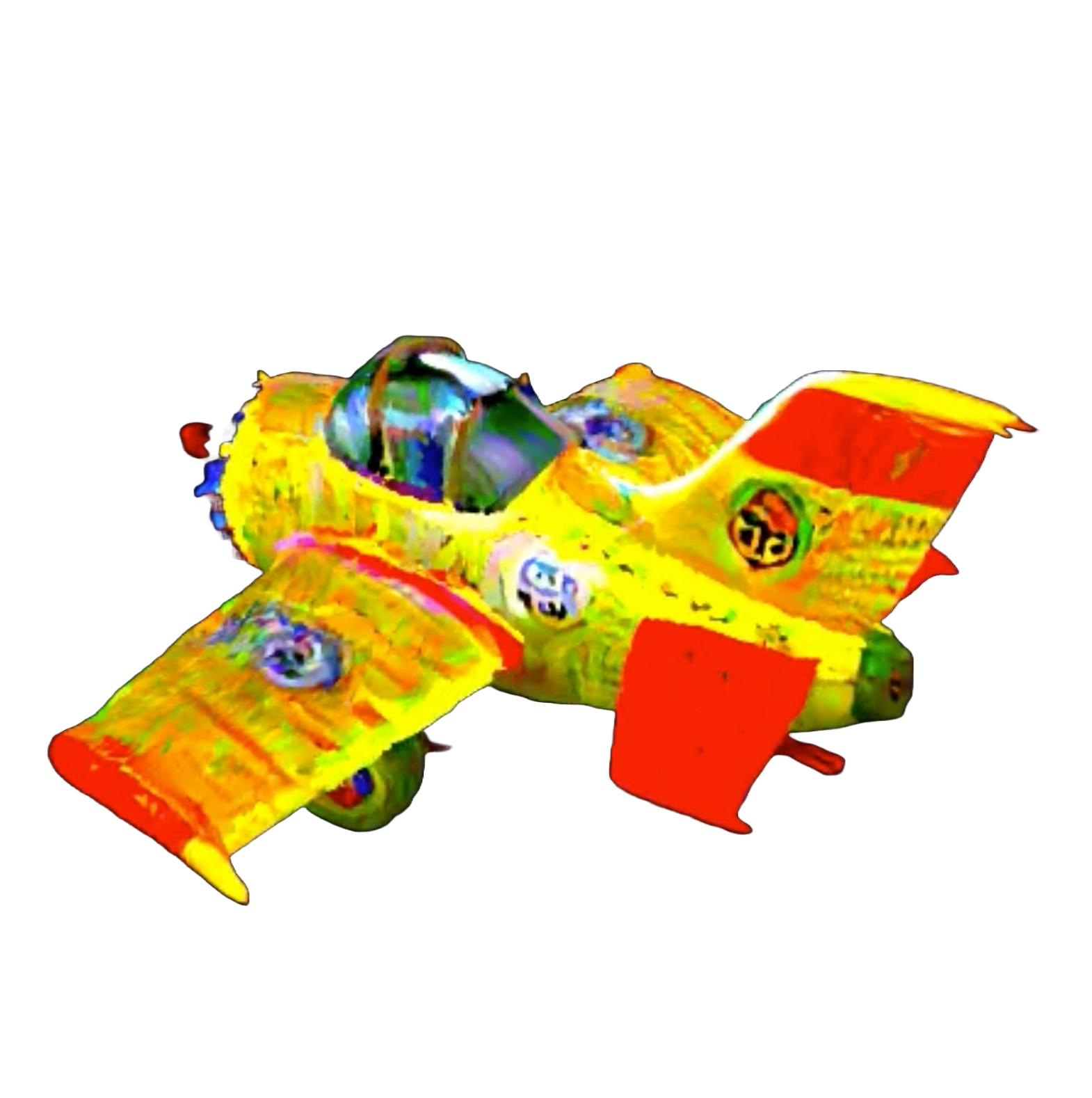} \\
        \multicolumn{8}{c}{"An opulent couch from the palace of Versailles"} \\
        \includegraphics[width=0.1\linewidth]{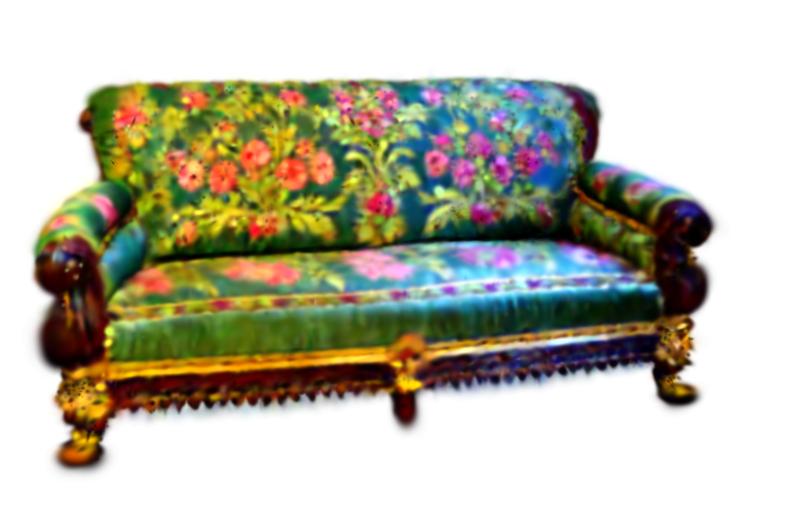} & 
        \includegraphics[width=0.1\linewidth]{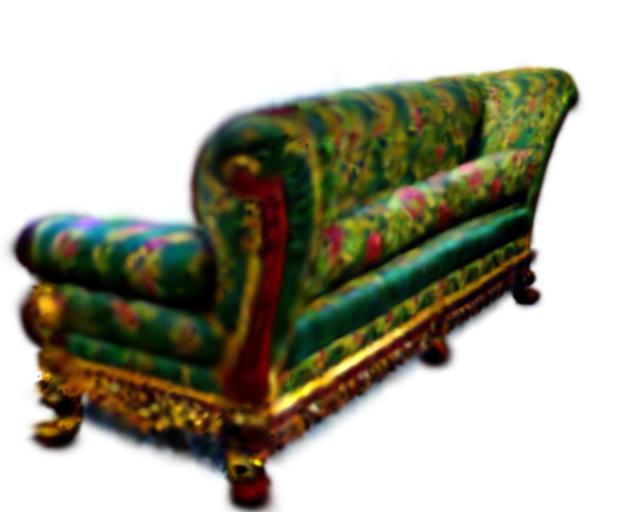} & 
        \includegraphics[width=0.13\linewidth]{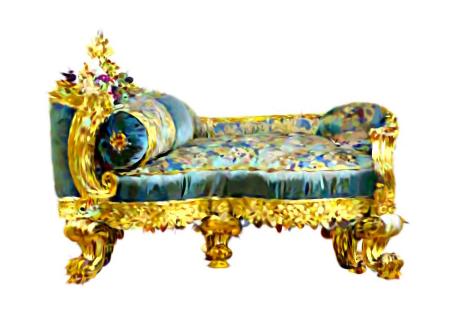} &
        \includegraphics[width=0.13\linewidth]{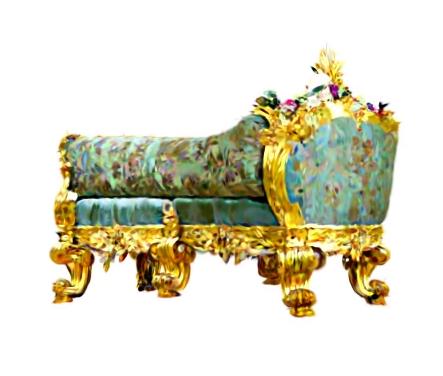} &
        \includegraphics[width=0.12\linewidth]{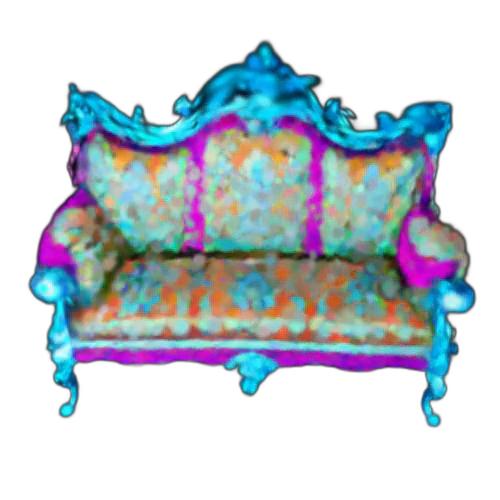} &
        \includegraphics[width=0.12\linewidth]{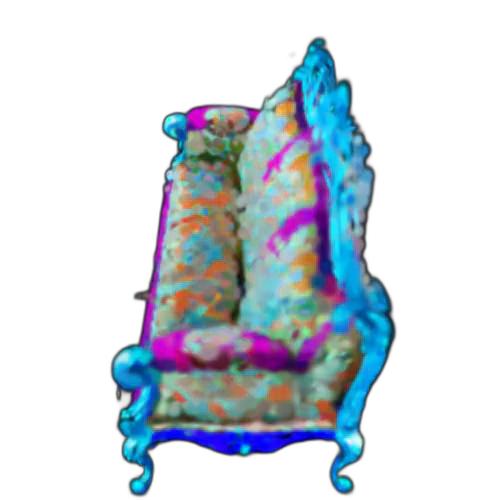} & 
        \includegraphics[width=0.14\linewidth]{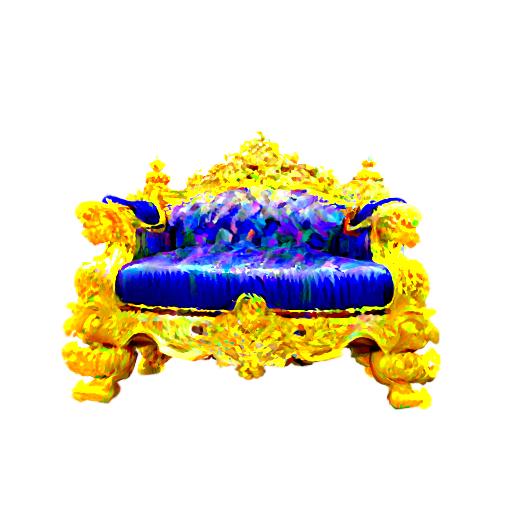} &
        \includegraphics[width=0.14\linewidth]{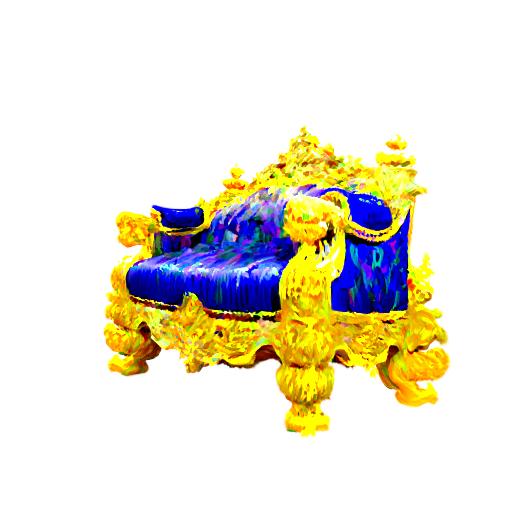} \\
        \multicolumn{8}{c}{"A portrait of Hatsune Miku as a robot, head, anime, super detailed, best quality, 8K, HD"} \\
        \includegraphics[width=0.10\linewidth]{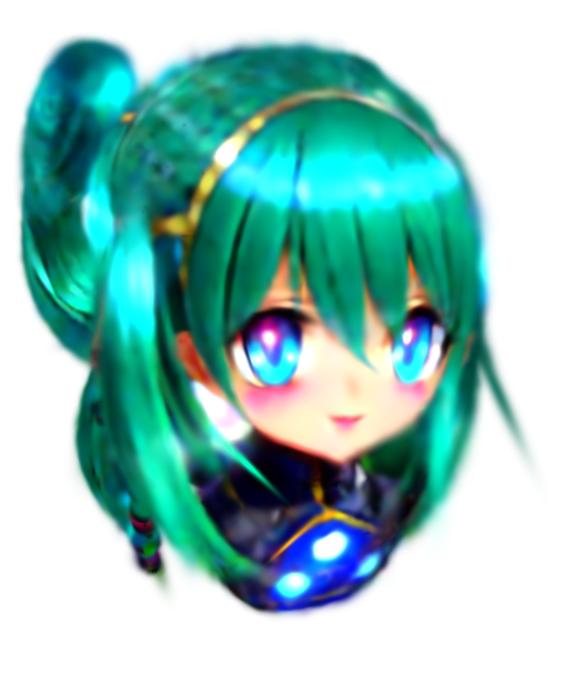} &  
        \includegraphics[width=0.10\linewidth]{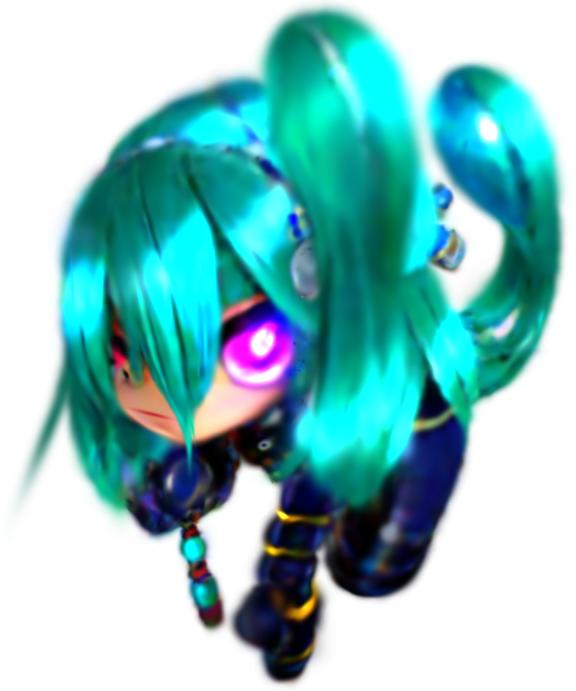} & 
        \includegraphics[width=0.12\linewidth]{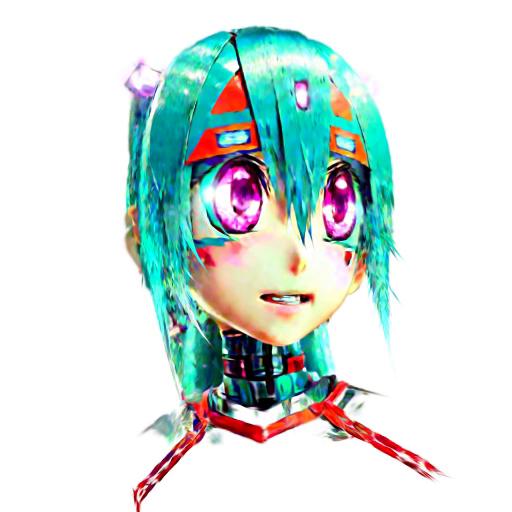} &
        \includegraphics[width=0.12\linewidth]{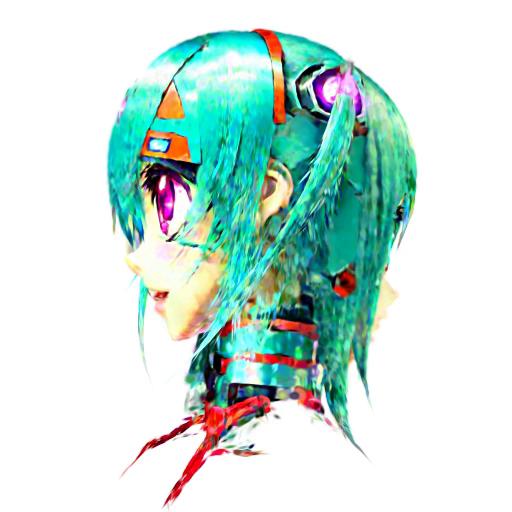} & 
        \includegraphics[width=0.10\linewidth]{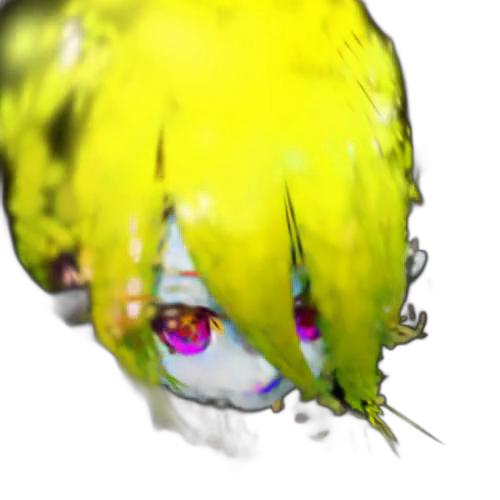} &  
        \includegraphics[width=0.1\linewidth]{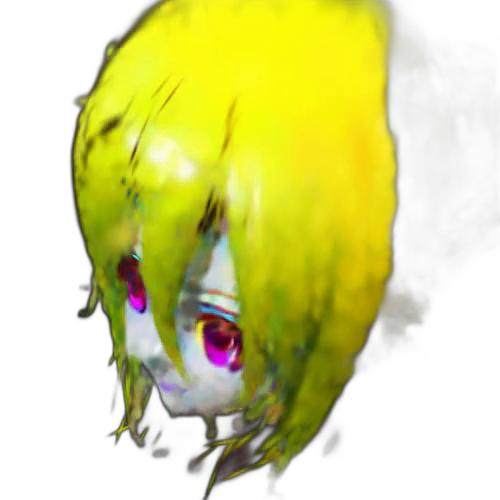} &
        \includegraphics[width=0.12\linewidth]{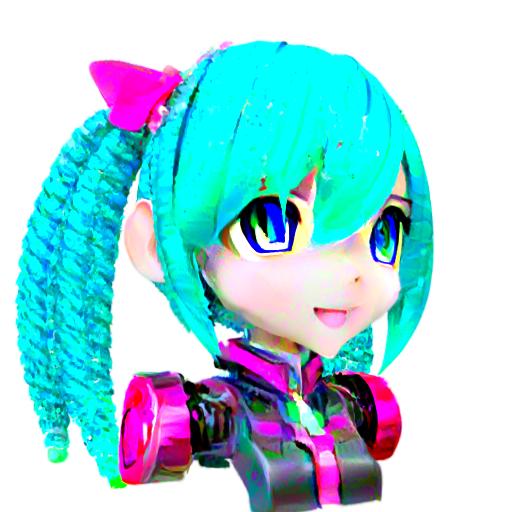} &
        \includegraphics[width=0.12\linewidth]{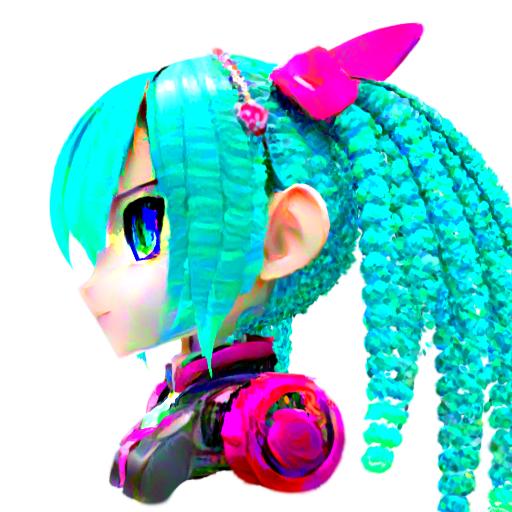} \\
        \multicolumn{8}{c}{"Flamethrower, with fire, scifi, cyberpunk, photorealistic, 8K, HD"} \\
        \includegraphics[width=0.13\linewidth]{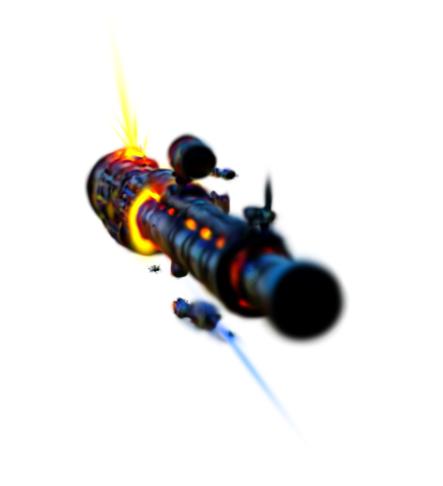} &  
        \includegraphics[width=0.12\linewidth]{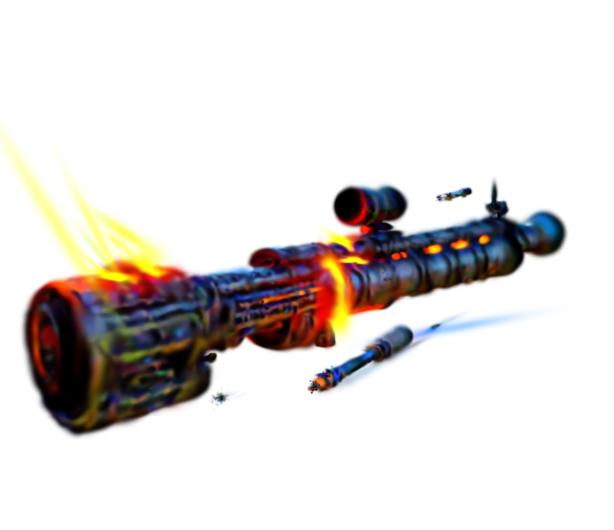} &
        \includegraphics[width=0.14\linewidth]{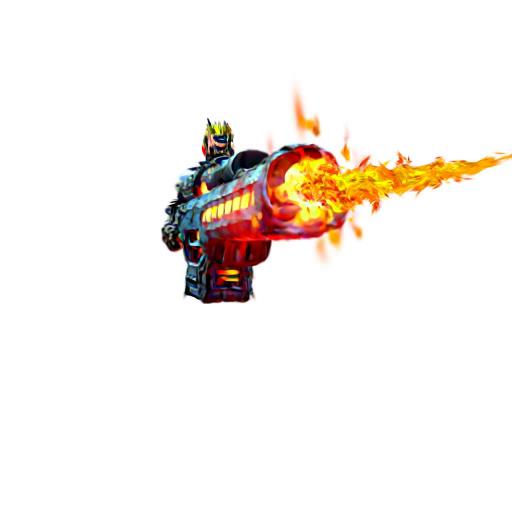} &
        \includegraphics[width=0.14\linewidth]{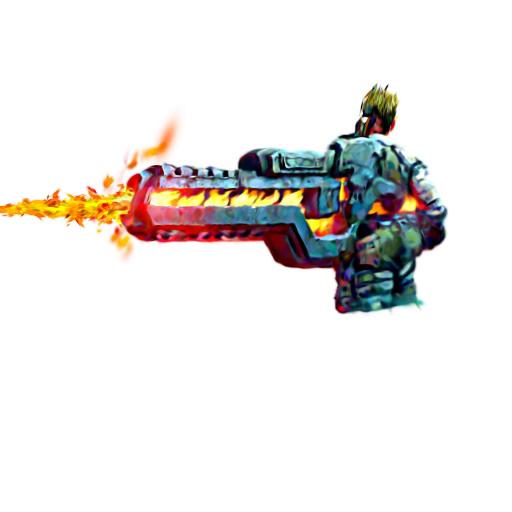} & 
        \includegraphics[width=0.1\linewidth]{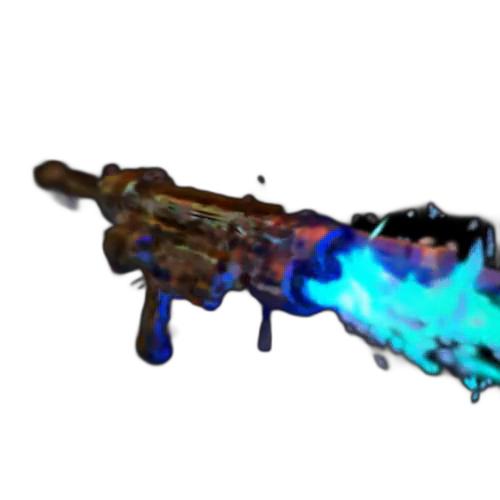} &  
        \includegraphics[width=0.1\linewidth]{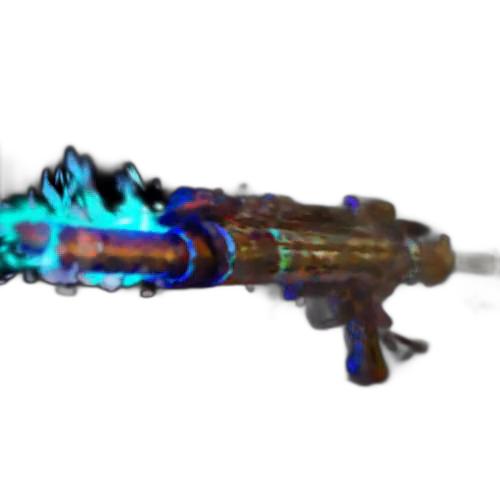} &
        \includegraphics[width=0.13\linewidth]{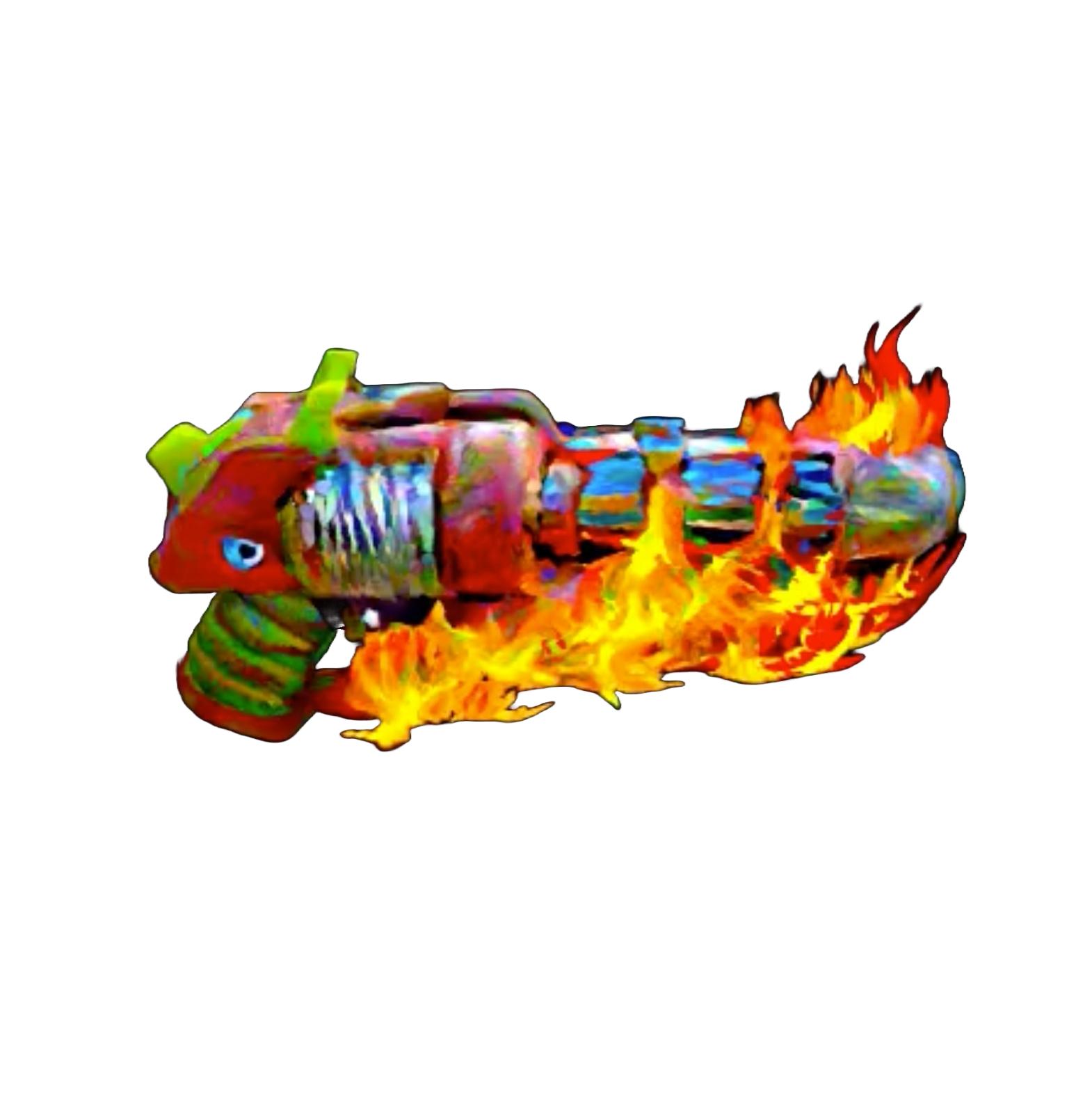} &
        \includegraphics[width=0.13\linewidth]{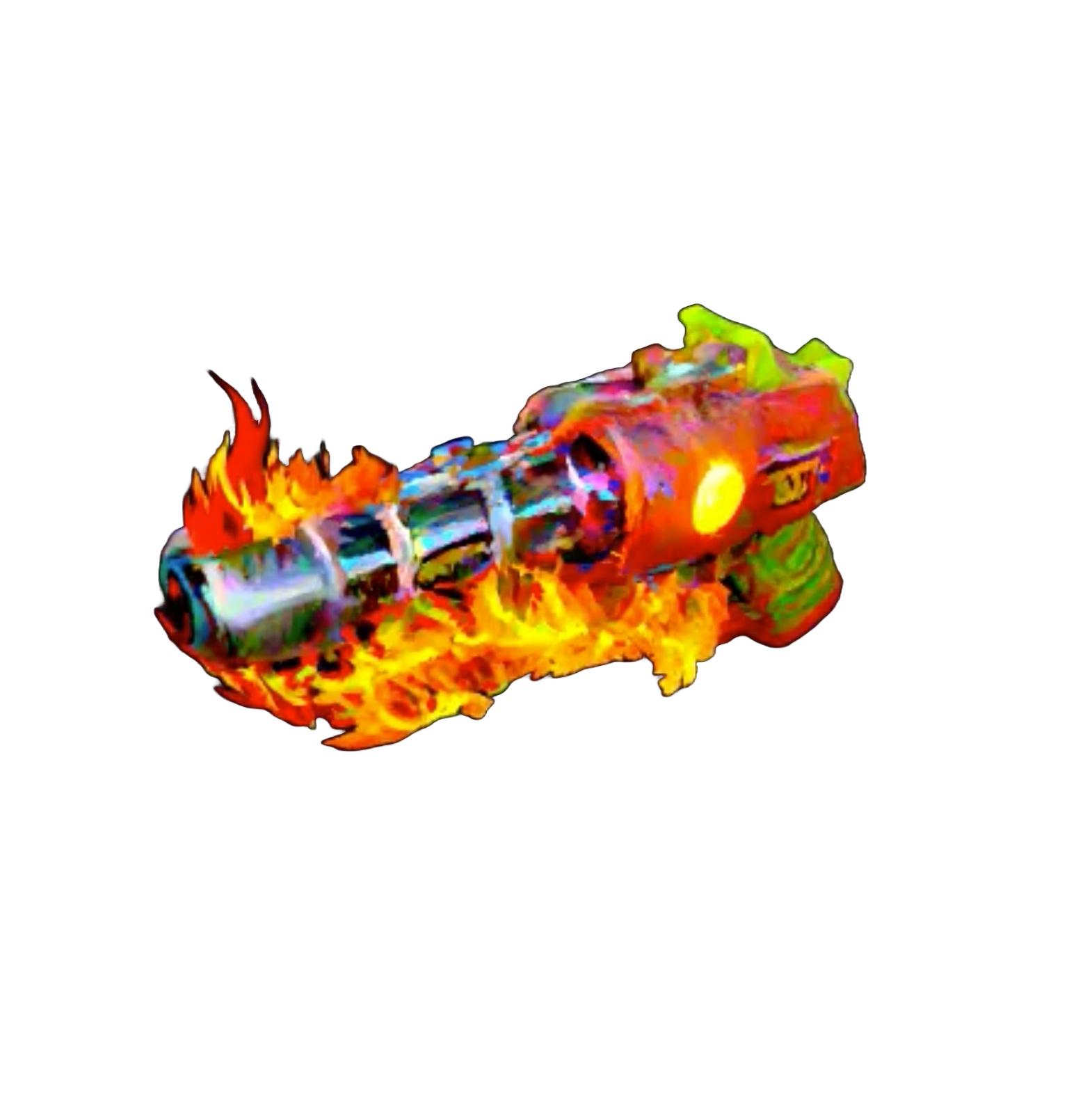}
    \end{tabular}}
    \caption{Additional comparison results with existing methods.}
    \label{fig:supp_comparisons}
\end{figure*}

Figure \ref{fig:supp_comparisons} presents additional qualitative comparisons of our method, MVGaussian, with GaussianDreamer, LucidDreamer, and GSGEN across various prompts. 

For the \textbf{\textit{Michelangelo dog statue}}, our model accurately captures both the style and the cellphone, while GaussianDreamer and GSGEN miss the cellphone, and LucidDreamer suffers from a multi-face Janus issue. In the \textbf{\textit{Steampunk airplane}} prompt, our method integrates the steampunk aesthetic effectively, unlike other methods that produce fighter jets without the steampunk elements. For the \textbf{\textit{Opulent couch}} prompt, GaussianDreamer, LucidDreamer, and MVGaussian produce detailed, prompt-aligned models, whereas GSGEN’s output is of lower quality. In the \textbf{\textit{Hatsune Miku robot}} prompt, our method captures the anime aesthetics, avoiding the distortions seen in other methods. Finally, in the \textbf{\textit{Flamethrower}} prompt, MVGaussian, along with GaussianDreamer and LucidDreamer, produces detailed and cohesive models, while GSGEN’s result lacks artistic detail and quality.

Overall, MVGaussian outperforms other methods in producing high-quality, detailed, and prompt-aligned 3D models.

\subsection{Implementation Details}

% Our implementation was built upon the DreamGaussian framework \cite{dreamgaussian}, maintaining a similar code structure. However, unlike DreamGaussian, our approach features a single training stage without a refining phase, incorporating several significant modifications.

\textbf{3D Gaussian Splatting (3DGS):} Similar to the original 3DGS implementation \cite{3DGS}, we primarily retain the initial learning rates for positions, color, opacity, scaling and rotation, with adjustments to clip these values to prevent excessively small rates that could hinder convergence. The learning rate ranges are [$1.6\times 10^{-4}, 1.6 \times 10^{-6}$] for position, [$3\times 10^{-3}, 2.5 \times 10^{-3}$] for color, [$0.1, 0.05$] for opacity, [$5\times 10^{-3}, 1 \times 10^{-3}$] for scaling, and [$1\times 10^{-3}, 2 \times 10^{-4}$] for rotation. We initialize $5000$ Gaussians and train for $10000$ iterations, initiating the densification and pruning process after $1000$ iterations and repeating it every $200$ iterations. We avoid starting densification too early or too frequently, as this can lead to the creation of redundant Gaussians that complicate optimization. Densification and pruning are halted after $8000$ iterations, allowing the final $2000$ iterations to focus on optimizing the existing Gaussians.

Densification is performed by cloning or splitting Gaussians with accumulated gradients greater than 0.05, while pruning removes Gaussians with opacity below 0.05. Additionally, we apply a surface pruning technique to eliminate redundant Gaussians, retaining only those near the surface. Figure \ref{code:pruning} provides pseudocode for our pruning algorithm. For each surface point, we identify the $5$ nearest neighbors from the set of Gaussian centers, preserving these close-to-surface Gaussians while pruning those farther away. We also offer an option to retain a percentage \(p\) of the remaining Gaussians by calculating the distances from Gaussian centers to the surface and pruning those with distances exceeding the threshold determined by the \(p\) percentile.

% \begin{lstlisting}[language=Python]
% \begin{figure*}[!htp]
% \centering
% \small
% \begin{minted}[frame=lines, linenos, breaklines]{python}
    \begin{figure*}[h]

\begin{lstlisting}[language=Python,numbers=none]

def get_surface_points(gaussians):
    all_points = list()
    azimuths, elevations = sample_camera_positions()
    
    for (azimuth, elevation) in zip(azimuths, elevations):
        camera_pose = get_camera_pose(azimuth, elevation)
        image, depth = render_views_from_3dgs(camera_pose, gaussians)
        R, T = get_cam_parameters()
        points_xyz = project_image2world(image, depth, R, T)
        points_xyz = remove_low_density_points(points_xyz)
    all_points = add_points(points_xyz)
    
    return all_points

def surface_pruning(percentile=0):
    # Get surface points by backprojecting points from image space to world space
    surface_points = get_surface_points(gaussians)
    gaussian_centers = get_gaussian_centers()

    if percentile > 0:
        dists = compute_knn_distances(gaussian_centers, surface_points, neighbors=1)
        dist_thresh = compute_quantile(dists, dist_percentile=percentile)
        # Filter out points that are too far from the surface point cloud
        prune_mask = dists > dist_thresh
    else:
        idxs = compute_knn_indices(surface_points, gaussian_centers, neighbors=5)
        prune_mask = compute_pruning_mask(idxs)
    remaining_gaussians = prune_gaussians(gaussian_centers, prune_mask)

    return remaining_gaussians

 \end{lstlisting}
\caption{Surface Point Pruning Pseudocode}
\label{code:pruning}
\end{figure*}
% \end{lstlisting}

\textbf{Multi-view Guidance:} We utilize the pretrained model (\textit{sd-v2.1-base-4view}) provided by MVDream \cite{mvdream}, which is based on the diffusion checkpoint at \textit{stabilityai/stable-diffusion-2-1-base}. To adapt MVDream's guidance, we generate four views that are linearly distributed around the object at a random elevation. These four views are then used to compute the SDS loss as introduced in DreamFusion \cite{dreamfusion}. We also use the following negative prompt to guide the generation model: \textit{"shadow, oversaturated, low quality, unrealistic, ugly, bad anatomy, blurry, pixelated, obscure, unnatural colors, poor lighting, dull, unclear, cropped, lowres, low quality, artifacts, duplicate, morbid, mutilated, poorly drawn face, deformed, dehydrated, bad proportions."}

The overall algorithm of our approach is outlined in Algorithm \ref{alg:final}. It involves iteratively optimizing Gaussian parameters with integrated densification and pruning. The process includes rendering multiple views, computing losses, and refining the model by selectively densify and prune existing Gaussians.
\\
\textbf{Hardware Setup:} Our experiments were run on a system equipped with an NVIDIA A100 Tensor Core GPU with 80 GB of VRAM, supported by two AMD EPYC 7543 32-Core Processors. The system supports a total of 128 CPUs.

\begin{algorithm}[H]
\caption{Overall Optimization Process}
\small
\begin{algorithmic}[1]
\State Initialize optimizers for Gaussian parameters and loss weights.
\For{each $i$ from 1 to $N$}
    \State Randomize 4 camera positions
    \State Randomly set background color
    \State Render RGB and depth from the 4 positions
    \State Compute SDS loss
    \State Compute regularization losses
    \State Compute final weighted loss
    \State Perform backpropagation
    \If {densification is required}
        \State Compute surface point cloud
        \State Perform surface pruning
        \If {distance $>$ threshold}
            \State Remove Gaussians
        \EndIf

        \If {opacity reset is required}
            \State Reset opacity
        \EndIf
    \EndIf
    \State Perform optimization steps
\EndFor
\end{algorithmic}
\label{alg:final}
\end{algorithm}

\subsection{Limitations \& Failure Cases}

\begin{figure}[!htp]
    \centering
    \scalebox{0.85}{
    \begin{tabular}{cc}
    "portrait of iron man  "      &   
    "warrior on a horse"
    \\
         \includegraphics[width=0.45\linewidth]{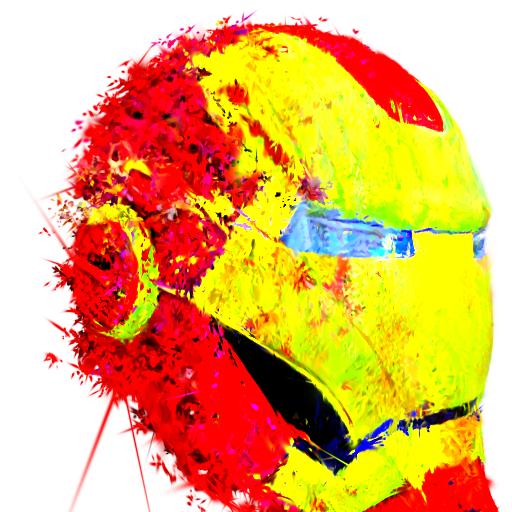} &
         \includegraphics[width=0.45\linewidth]{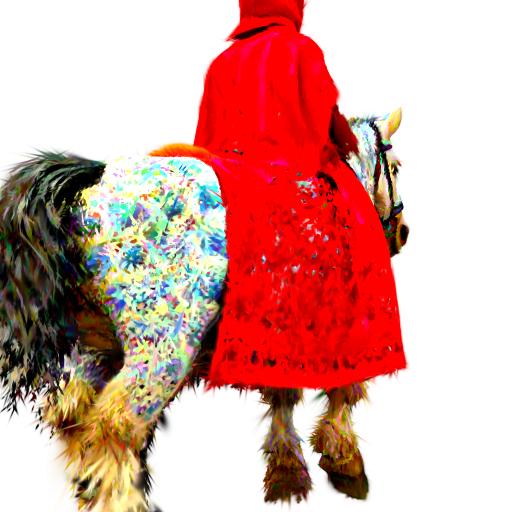} \\
         "a DSLR photo of eiffel tower"
         &
         "A white plate piled high \\
         &  with chocolate chip cookies" \\
         \includegraphics[width=0.45\linewidth]{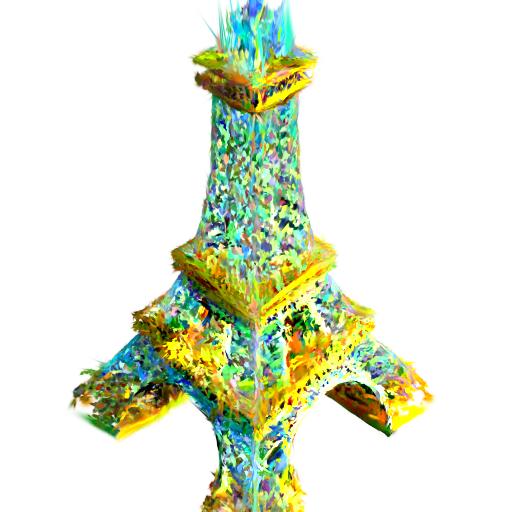} &
         \includegraphics[width=0.45\linewidth]{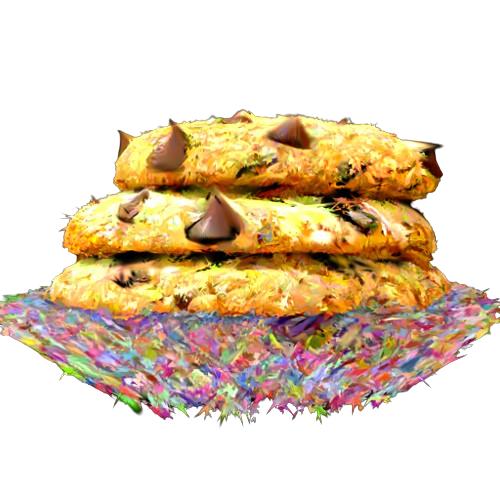} 
         
    \end{tabular}}
    \caption{Failure Cases, usually contain spike-like artifacts due to lack on convergence or over colorization like in the plate containing cookies and the horse in row 1.}
    \label{fig:failure_cases}
\end{figure}

The failure cases shown in Figure \ref{fig:failure_cases} highlight issues specific to our approach, particularly the occurrence of spiky Gaussian artifacts and over-colorization. The spiky artifacts, which are noticeable in models such as the \textbf{\textit{plat of cookies}} and the \textbf{\textit{Eiffel Tower}} could be a consequence of the flatness regularization used in our method. This regularization may inadvertently introduce sharp spikes on the surface of the models, disrupting their smoothness. Additionally, over-colorization is evident in the \textbf{\textit{Iron Man}} and the \textbf{\textit{warrior on a horse}}, where the colors are overly saturated and do not align with the intended realistic appearance. These failure cases underscore the challenges in balancing regularization techniques to avoid such artifacts while maintaining high visual quality in the generated 3D models.

These issues could be mitigated with longer training, allowing the model to better converge and reduce such artifacts. Despite these occasional imperfections, our method tends to perform better across a wider range of prompts compared to other methods, demonstrating its overall robustness and effectiveness in generating high-quality 3D content.

Another limitation is the evaluation criteria. Despite notable visual improvements, the CLIP score does not accurately reflect the quality of the generated content. For example, in \textbf{Figure 2 of the main manuscript}, the \textbf{\textit{Castle}} appears more photorealistic and highly detailed, yet it receives a lower CLIP score of $0.30$ as compared to a score of $0.32$ achieved by GaussianDreamer. Similarly, for the \textbf{\textit{Green Orc}}, GSGEN, despite exhibiting the multi-face Janus problem, achieves the same score as our method with $0.31$. This inconsistency highlights the unreliability of the CLIP score in evaluating 3D model quality.

Additionally, we observe that directly using 2D Gaussian Splatting (2DGS) as a densification strategy introduces artifacts, as illustrated in Figure \ref{fig:2dvs3d}. Further investigation into the incorporation of 2DGS is left for future research.

% Spike-like artifacts are observed in several samples due to convergence issues as shown in Fig. \ref{fig:failure_cases}. 
% \vspace{-10pt}

\subsection{Additional Results}

\begin{figure*}[htp]
    \includegraphics[height=1.0\textheight]{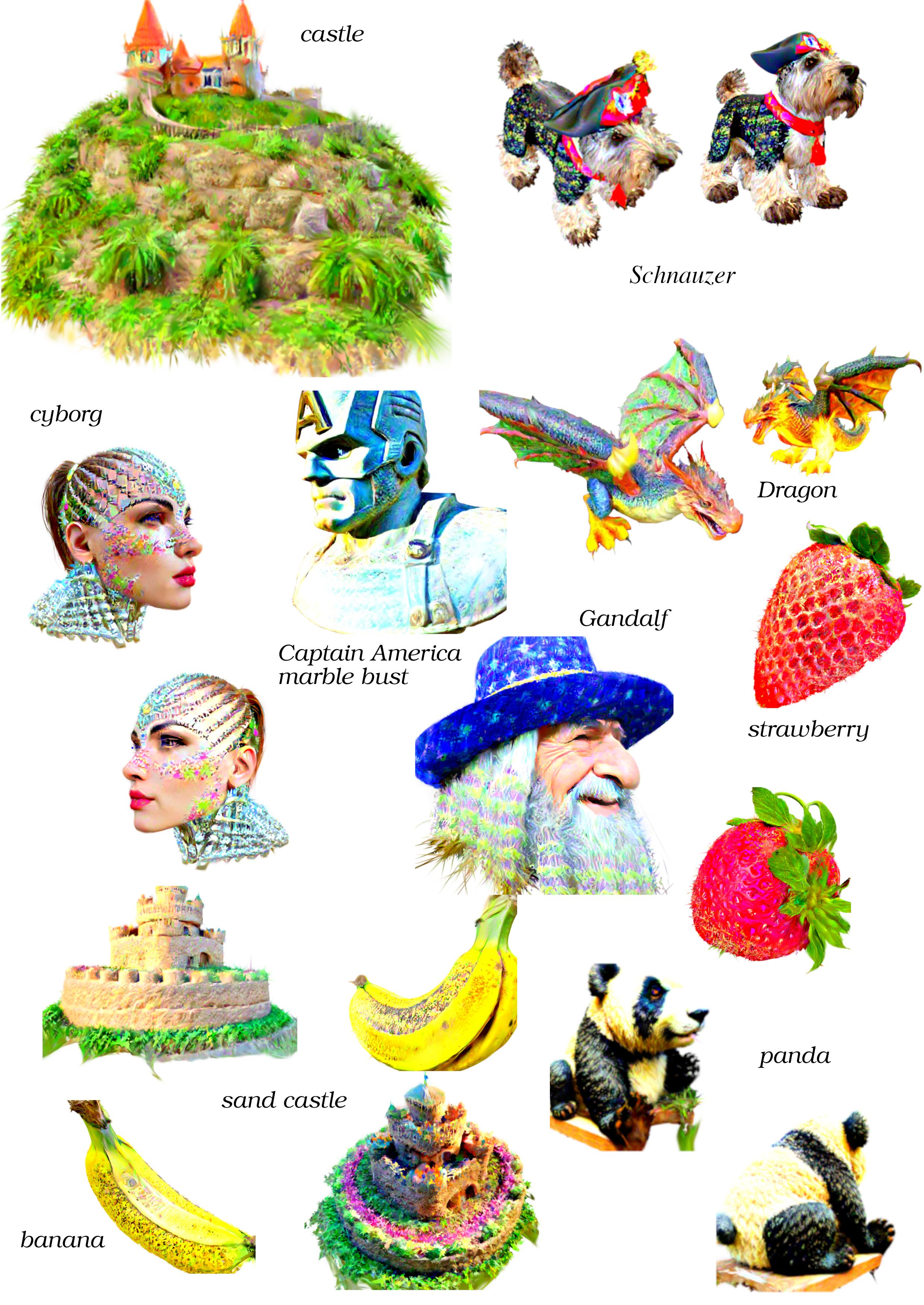}
    \caption{Additional results generated using generic prompt template such "a DSLR photo of a ..." }
    \label{fig:additional_res}
\end{figure*}

Figure \ref{fig:additional_res} showcases additional results generated by our model using a generic prompt template such as \textbf{\textit{``a DSLR photo of a ..."}}. The diverse range of objects includes a castle, a cyborg, a marble bust of Captain America, a dragon, Gandalf, and so on. These examples demonstrate the model's capability to create highly detailed and visually coherent 3D representations across various subjects, from complex fantasy characters to everyday objects, illustrating the robustness and versatility of our approach.

\textbf{Result videos:} Due to file size constraints, we cannot include all the videos of the generated models in this paper. Please check our project page at https://mvgaussian.github.io for the video results of the generated content reported in this paper.

\end{document}